\documentclass[]{fairmeta}

\usepackage{wrapfig}
\usepackage{tabularx}
\usepackage{textcomp}
\usepackage{stfloats}
\usepackage{url}
\usepackage{verbatim}
\usepackage{titlesec}
\usepackage{tocloft}
\usepackage{adjustbox}
\usepackage{multirow}
\usepackage{pifont}
\usepackage{tikz}
\usepackage{comment}
\usepackage{amsmath,amssymb} 
\usepackage{colortbl}  
\usepackage{color}
\usepackage{booktabs} 
\usepackage{hyperref}
\usepackage{graphicx}    
\usepackage{subcaption}
\usepackage{multirow} 
\usepackage{booktabs} 
\usepackage{subcaption} 

\usepackage[utf8]{inputenc} 
\usepackage{caption}
\usepackage[T1]{fontenc}    
\usepackage{amsfonts}       
\usepackage{nicefrac}       
\usepackage{microtype}      
\usepackage{xcolor}         
\usepackage{graphicx} 
\usepackage{amsmath}        
\usepackage{rotating}       
\usepackage{colortbl}       
\usepackage{wrapfig}
\usepackage{float}             
\usepackage[most]{tcolorbox} 
\usepackage{listings}        
\RequirePackage{xspace}
\makeatletter
\DeclareRobustCommand\onedot{\futurelet\@let@token\@onedot}
\def\@onedot{\ifx\@let@token.\else.\null\fi\xspace}

\makeatother

\definecolor{adptorange}{RGB}{248, 205, 172}
\definecolor{cmpblue}{RGB}{189, 215, 238}
\definecolor{cmpblue}{RGB}{189, 215, 238}

\definecolor{our_red}{RGB}{232,157,160}
\definecolor{our_blue}{RGB}{136,206,230}
\definecolor{our_orange}{RGB}{246,200,168}
\definecolor{our_green}{RGB}{178,211,164}

\definecolor{attn_code0}{RGB}{247,215,200}
\definecolor{attn_code1}{RGB}{238,169,139}
\definecolor{mlp_code0}{RGB}{204,201,221}
\definecolor{mlp_code1}{RGB}{102,95,153}

\definecolor{token_blue}{RGB}{84, 120, 140}

\usepackage{pifont}       
\usepackage{bbding}       
\usepackage{fontawesome}
\usepackage{xspace}

\usepackage{float}

\newlength\savewidth

\newcolumntype{x}[1]{>{\centering\arraybackslash}p{#1pt}}
\newcolumntype{y}[1]{>{\raggedright\arraybackslash}p{#1pt}}
\newcolumntype{z}[1]{>{\raggedleft\arraybackslash}p{#1pt}}

\renewcommand{\paragraph}[1]{\vspace{1mm}\noindent\textbf{#1}}
\usepackage{colortbl}
\usepackage{xcolor}
\usepackage{wrapfig}

\renewcommand{\paragraph}[1]{\vspace{1.25mm}\noindent\textbf{#1}}

\usepackage{algorithm}
\usepackage{listings}

\definecolor{codeblue}{rgb}{0.25, 0.5, 0.5}
\definecolor{codekw}{rgb}{0.35, 0.35, 0.75}
\lstdefinestyle{Pytorch}{
    language = Python,
    backgroundcolor = \color{white},
    basicstyle = \fontsize{9pt}{8pt}\selectfont\ttfamily\bfseries,
    columns = fullflexible,
    aboveskip=1pt,
    belowskip=1pt,
    breaklines = true,
    captionpos = b,
    commentstyle = \color{codeblue},
    keywordstyle = \color{codekw},
}

\definecolor{green}{HTML}{009000}
\definecolor{red}{HTML}{ea4335}

\title{BreastGPT: A Multimodal Large Language Model for the Full Spectrum of Breast Cancer Clinical Routine}

\author[* 1]{Yang Liu}
\author[* 1, 3]{Jiajin Zhang}
\author[* \dagger 1, 3]{Danyang Tu}
\author[2]{Yaojun Hu}
\author[4]{Jiao Qu}
\author[5]{Jiuyu Zhang}
\author[5]{Yu Shi}
\author[* 1, 3]{Wei Fang}
\author[\dagger 2]{Shi Gu}
\author[1]{Ling Zhang}
\author[1]{Yingda Xia}

\affiliation[1]{DAMO Academy, Alibaba Group}
\affiliation[2]{Zhejiang University\\}
\affiliation[3]{Hupan Lab}
\affiliation[4]{West China Hospital}
\affiliation[5]{China Medical University}
\contribution[*]{Equal contribution}
\contribution[\dagger]{Corresponding author}

\abstract{
Breast cancer remains a leading cause of cancer-related mortality among women. Its clinical management requires multimodal reasoning across a clinical workflow that spans \textit{screening}, \textit{diagnosis} and \textit{treatment planning}, where each stage involves distinct imaging modalities, task objectives, and reasoning patterns. However, constrained by data scarcity and model versatility, existing medical MLLMs are typically evaluated on isolated modalities or narrow task families, limiting their ability to support workflow-level clinical reasoning. In this work, we first introduce \textbf{BreastStage}, a workflow-aligned breast imaging instruction corpus comprising 1.86M instruction-following pairs curated from 17 sub-datasets across 5 imaging modalities and 136 task templates. Its held-out split, \textbf{BreastStage-Bench}, provides a comprehensive benchmark for evaluating multimodal reasoning across the breast cancer care continuum. Building on this corpus, we propose \textbf{BreastGPT}, a unified MLLM equipped with a dual-branch visual encoder and concept-preserving token compression to bridge the scale gap between standard radiology and gigapixel pathology. On BreastStage-Bench, BreastGPT achieves 75.66\% closed-ended accuracy and 89.92\% open-ended score, outperforming both general-purpose and medical-specific MLLMs across clinical stages and task formats. These results suggest that workflow-aligned data and cross-scale visual modeling are critical for clinically grounded medical MLLMs. All data, code, and model checkpoints are released at \url{https://yangyy-liu.github.io/BreastGPT.io}.
}

\date{\today}

\begin{document}
\thispagestyle{firstheader}
\maketitle
\addtocontents{toc}{\protect\setcounter{tocdepth}{-1}}
\pagestyle{empty}

\section{Introduction} \label{sec:introduction}
Breast cancer remains one of the most prevalent and life-threatening malignancies among women worldwide~\cite{anon2025current,anon2025global,anon2025globalb,anon2025globalc}. Its clinical management involves not just a single diagnostic decision, but a staged workflow spanning \textit{screening}, \textit{diagnosis}, and \textit{treatment planning}. Each stage relies on distinct imaging evidence and pursues different clinical objectives: \textit{Screening} prioritizes triage at the population level, where mammography, breast ultrasound (BUS) and chest CT are mainly used to detect early suspicious findings and estimate their risk~\cite{magnenat2024screening}; \textit{Diagnosis} then shifts toward fine-grained lesion characterization, where multimodal MRI is pivotal for evaluating morphology and probability of malignancy~\cite{hadadi2022comparison}; \textit{Treatment planning} requires tumor biology and response assessment, for which gigapixel whole-slide images (WSIs) provide pathological subtyping and biomarker evidence~\cite{krishnamurthy2024ai}. This clinical continuum imposes a different requirement from conventional single-task image understanding: a clinically useful model must process heterogeneous modalities while preserving the stage-specific reasoning style expected at each point of care.

However, while recent advances in MLLMs have significantly expanded the scope of medical image understanding~\cite{anon2025visuallanguage,anon2025generalpurpose,anon2025generalist,anon2025comprehensive,anon2025multimodal,anon2025vision,anon2025multimodalb,chen2024huatuogpt,lin2025healthgpt}, 
recent efforts in breast imaging AI are mainly tailored for individual modalities or narrowly defined tasks. Specifically, mammography-oriented models~\cite{versamammo2025,mammovqa2025} achieve strong performance on mammograph interpretation; BUS-CoT~\cite{buscot2025} introduces structured reasoning for breast ultrasound images, and MOME~\cite{mome2024} supports multimodal MRI analysis with performance comparable to experienced radiologists. Despite demonstrating promising modality-specific performance, these methods conventionally inherit a fragmented formulation of breast cancer AI: datasets, models, and benchmarks are invariably organized around isolated imaging modalities and limited tasks for a single clinical stage. Consequently, 1) these single-modality datasets fail to comprehensively evaluate models' performance across the various stages within the practical clinical workflow; 2) there is a lack of a versatile model capable of processing different modalities uniformly and efficiently to address various tasks at different stages of real-world clinical workflows.

\begin{figure}[t]
  \centering
  \includegraphics[width=\linewidth]{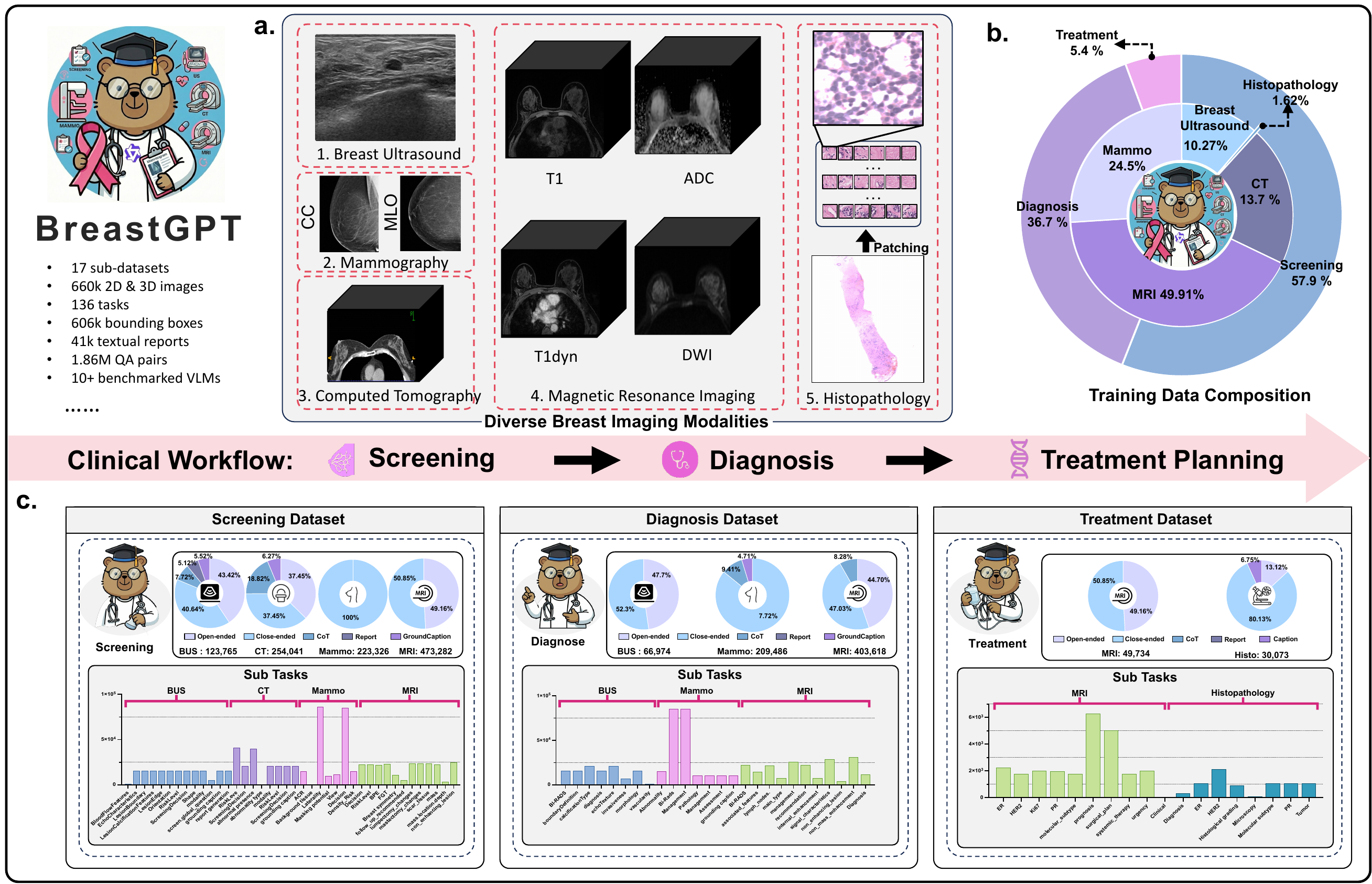}
  \caption{Overview of the \textbf{BreastStage} dataset aligned with the end-to-end breast cancer clinical workflow. The dataset is structured across three clinical stages: (1) \textbf{Screening} utilizes mammography and breast ultrasound (BUS) images for early lesion detection; (2) \textbf{Diagnosis} leverages MRI (main modality), BUS and Mammography for detailed tumor characterization; and (3) \textbf{Treatment planning} involves gigapixel whole-slide images (WSI) for pathological subtyping and therapeutic evaluation.}
  \label{fig:BreastStage}
\end{figure}

To this end, we first construct BreastStage (Fig. \ref{fig:BreastStage}), a large-scale multimodal instruction corpus aligned with the breast cancer clinical workflow. BreastStage integrates \textbf{17} sub-datasets across \textbf{5} modalities (mammography, BUS, MRI, WSI, and CT), yielding \textbf{1.86M} instruction-following QA pairs over \textbf{136} task templates. These instructions cover attribute extraction, report generation, visual question answering, and image-grounded captioning, providing supervision over both radiological and pathological evidence.  Based on the held-out split of BreastStage, we further build \textbf{BreastStage-Bench}, a comprehensive benchmark for workflow-level breast oncology reasoning developed with breast oncology experts. We assess multiple publicly available multimodal large language models, including both general-purpose and medical-specific variants, alongside advanced proprietary models. Extensive experiments demonstrate that BreastStage-Bench poses a significant challenge for current MLLMs: GPT-5.4 achieves an average score of only 49.32\%, and existing medical MLLMs show no clear advantage over general-purpose models on these clinical workflow-aligned tasks.

Based on BrasetStage, we propose \textbf{BreastGPT}, a unified MLLM for breast cancer workflow reasoning. Built upon Qwen3-VL~\cite{qwen35vl2025}, BreastGPT handles all five imaging modalities within a single architecture and adopts stage-conditioned system prompts to switch between screening, diagnostic, and treatment-oriented reasoning behaviors. Meanwhile, to address the significant differences in image scale across modalities, BreastGPT adopts a dual-branch visual encoder with a resolution-aware gating module that automatically routes each input to an appropriate processing branch based on stage condition. Aligned with practical clinical workflow, BreastGPT demonstrates exceptional versatility and achieves remarkable performance across various tasks. Compared to existing general-purpose and medical-specific MLLMs, BreastGPT achieves performance gains of over 25\%, 35\%, and 40\% for screening, diagnosis, and treatment planning, respectively. To summarize, our contributions are threefold:

\noindent\textbf{Workflow-level problem formulation and benchmark.}
We formulate breast cancer multimodal reasoning as a workflow-aligned problem spanning screening, diagnosis, and treatment planning. Based on this, we construct BreastStage, a large-scale instruction corpus with 1.86M QA pairs from 17 sub-datasets across 5 modalities, and BreastStage-Bench, a held-out benchmark for evaluating stage-specific clinical reasoning.

\noindent\textbf{Unified cross-scale breast MLLM.}
We introduce BreastGPT, a unified multimodal LLM that handles mammography, ultrasound, CT, MRI, and gigapixel pathology within a single instruction-following framework, enabling consistent reasoning across the full breast cancer clinical routine.

\noindent\textbf{Concept-preserving visual token compression.}
To bridge the extreme scale gap between standard radiology and WSIs, we propose a dual-branch visual encoder with a concept-based token selector that preserves clinically salient visual evidence under a fixed token budget, substantially improving both accuracy and inference efficiency.

\begin{figure}[!t]
  \centering
  \includegraphics[width=1\linewidth]{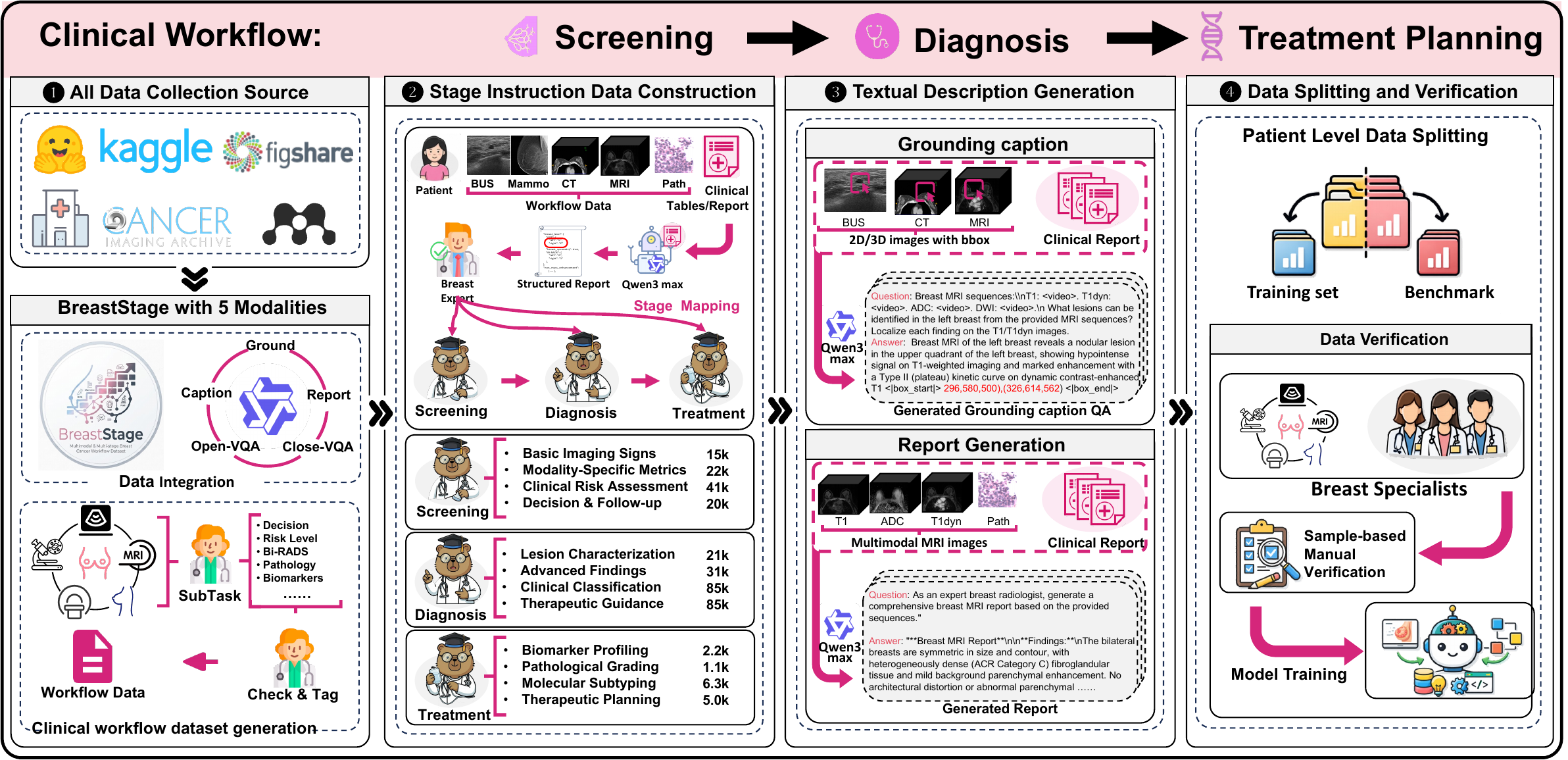}
  \caption{\textbf{BreastStage construction pipeline.} It involves curating diverse imaging modalities from public datasets and hospitals. The raw data is then cleaned, annotated, and checked by breast experts. Textual and instruction data are generated by prompting Qwen3-Max. Finally, the data is split into a training set and a benchmark, with professional breast experts assessing the training samples.}
  \label{fig:dataset pipeline}
\end{figure}

\section{Related Work}

\noindent\textbf{Medical multimodal large language models.}
Medical MLLMs extend general-purpose vision-language models to clinical image understanding, report generation, visual question answering, and interactive decision support~\cite{anon2025multimodal,anon2025vision,anon2025multimodalb,anon2025comprehensive,anon2025taming,anon2025visionlanguage}.
Representative systems, including LLaVA-Med, HuatuoGPT-Vision, MedDr, HealthGPT, and Lingshu, show that biomedical instruction tuning can improve clinical-domain perception and dialogue~\cite{li2023llavamed,chen2024huatuogpt,he2024meddr,lin2025healthgpt,anon2025generalist}.
In parallel, medical and general multimodal evaluation suites such as OmniMedVQA, GMAI-MMBench, MMMU, MMBench, and VLMEvalKit reveal persistent gaps between generic multimodal capability and reliable clinical reasoning~\cite{hu2024omnimedvqa,anon2025comprehensiveb,yue2024mmmu,liu2024mmbench,duan2024vlmevalkit}.
Recent specialty-specific studies in dentistry further suggest that clinically meaningful evaluation often requires benchmarks organized around anatomy, modality, reporting convention, and treatment workflow~\cite{hao2025dental,hao2025oralgpt}.
BreastGPT follows this specialty-grounded direction for breast oncology, where the central challenge is not only medical image perception but also workflow-aware reasoning across screening, diagnosis, pathology, and treatment.

\noindent\textbf{Breast imaging AI.}
Breast imaging AI has progressed rapidly, but most prior work remains modality-specific.
In mammography, VersaMammo builds a large multi-institutional mammogram foundation model and evaluates detection, segmentation, classification, retrieval, and VQA tasks~\cite{versamammo2025}, while MammoVQA standardizes mammogram VQA across public datasets and shows that both general-purpose and medical-specific LVLMs remain unreliable for mammogram interpretation~\cite{mammovqa2025}.
For breast ultrasound, BUS-CoT introduces reasoning annotations that connect ultrasound observations, imaging features, BI-RADS assessment, pathology labels, and clinical rationales~\cite{buscot2025}.
For breast MRI, MOME integrates multiparametric MRI through a mixture-of-modality-experts design and supports malignancy identification, biopsy recommendation, triple-negative breast cancer classification, and neoadjuvant chemotherapy response prediction~\cite{mome2024}.
These works demonstrate the value of breast-specific modeling, but they are primarily organized around individual modalities or task types rather than a unified breast-oncology workflow spanning screening, diagnosis, pathology, and treatment-related reasoning.

\noindent\textbf{Computational pathology and gigapixel slide modeling.}
Computational pathology has moved from patch-level representation learning toward slide-level and vision-language foundation models.
Pathology foundation models such as UNI, CONCH, and large-scale WSI encoders learn transferable morphology representations for diagnostic, prognostic, retrieval, and report-generation tasks~\cite{anon2025towardsb,anon2025visuallanguage,anon2025generalpurpose,anon2025millionslide}.
Meanwhile, WSI vision-language and long-sequence modeling approaches, including PathAlign, query-aware gigapixel modeling, hierarchical WSI transformers, and long-context histopathology transformers, seek to bridge local patch embeddings with slide-level reasoning and language interaction~\cite{anon2025visionb,anon2025queryaware,anon2025hierarchical,anon2025rethinking,anon2025survey}.
These studies highlight a key challenge for unified breast workflow modeling: WSIs contain sparse, distributed, and scale-dependent evidence, whereas radiology images are smaller, denser, and governed by different clinical semantics.
A model that jointly handles breast radiology and pathology must therefore reconcile substantially different visual scales and token budgets.

\noindent\textbf{Efficient visual token selection.}
Efficient visual token selection is essential for deploying MLLMs on high-resolution and long-context medical inputs.
Recent methods reduce inference cost through token pruning, learned token selection, adaptive compression, query-aware retrieval, and layer-wise token reduction~\cite{dong2025mmtok,anon2025adaptive,anon2025learning,anon2025effective,anon2025perlayer,anon2025token}.
MMTok formulates token selection as multimodal coverage maximization, balancing text-vision alignment with vision-vision diversity~\cite{dong2025mmtok}.
BreastGPT adapts this principle to breast workflow modeling by using a shared token budget across radiology and pathology inputs while preserving both prompt-relevant clinical evidence and global visual coverage.
This design allows gigapixel WSIs and conventional radiology images to be handled within a unified downstream language-model interface.

Existing work establishes strong foundations for medical MLLMs, breast imaging models, pathology foundation models, and efficient token compression, but these directions remain largely separate.
BreastGPT integrates them into a unified breast cancer workflow model, and BreastStage-Bench evaluates whether such integration supports stage-aware reasoning across heterogeneous modalities, clinical tasks, and care roles.
\section{BreastStage Dataset Curation}
\label{sec:dataset}
\subsection{Scope and scale}
BreastStage contains about 662{,}    000 unique 2D and 3D images, 136 task templates, 606{,}226 records with bounding-box or mask annotations, and 1.86 million instruction-following pairs. Task coverage spans classification, visual grounding, open- and closed-ended question answering, captioning, and report generation, so the dataset measures both perception accuracy on narrow subtasks and the model's ability to follow diverse clinical instructions. Samples are distributed as 57.9\% screening, 36.7\% diagnosis, and 5.4\% treatment, broadly tracking the relative prevalence and data availability of these stages in real breast care. Please check $\S$ \ref{supp:data_sources} for details.

\subsection{Data construction pipeline}
As illustrated in \cref{fig:dataset pipeline}, the BreastStage curation pipeline has four steps:

\noindent\textbf{i. Workflow Data Generation.} We assemble a cohort of breast imaging studies that draws CT, BUS, mammography, and WSI from public repositories and adds multimodal MRI from two collaborating clinical institutions. For quality control, modality-specific visual specialist agents built on Qwen2.5-VL-72B~\cite{bai2025qwen25vl} emit a structured \texttt{\{validity, reason\}} verdict for each image; samples flagged as low-quality are routed to a Low-Image Check in which a breast specialist confirms or restores the verdict, rather than being silently dropped. Lesion-level spatial annotations are then attached: BUS reuses the expert hand-drawn masks shipped with BUS-CoT~\cite{buscot2025}, mammography reuses the EMBED-released bounding boxes, CT volumes from CT-RATE~\cite{ctrate2024} are passed through DRT-M3D~\cite{zhou2026dualres} to obtain tumour segmentations, and MRI carries manual T1 and dynamic contrast-enhanced T1dyn annotations from a panel of 10 breast specialists. The resulting cohort is tagged with one of 136 specialist-designed sub-tasks (\texttt{task template}) and audited by hand. Full data provenance and per-modality counts are reported in Table~\ref{tab:source_summary} and the complete pipeline is detailed in $\S$ \ref{supp:instruction_pipeline}.

\noindent\textbf{ii. Stage Instruction Data Construction.} All data is standardized, cleaned, and paired with its matching clinical tables or reports, then converted into instruction-following QA pairs: \textit{1)} Chinese reports are first translated to English by Qwen3-Max acting as a radiologist agent; \textit{2)} Each translated report is then parsed into a modality-specific and BI-RADS-aligned structured report whose schema is designed by breast experts following clinical reporting guidelines; \textit{3)} For every \texttt{task template} from the previous stage, breast experts then write a closed-ended question whose answer is drawn from one or more leaf fields of the structured report, with options taken directly from the schema's enum values, so the resulting QA pairs cannot introduce facts beyond the structured record; \textit{4)} Finally, each \texttt{task template} is mapped, with expert guidance, to one of the three clinical stages: screening, diagnosis, or treatment.

After the stage mapping, the resulting BreastStage instruction set is organized to match the cognitive demands of each clinical stage, with category-level counts summarized in Table~\ref{tab:task_taxonomy_detailed}. Screening tasks emphasize early detection and triage, covering modality / view / laterality recognition, breast-tissue characterization (FGT, BPE, density), risk and decision tasks, lesion presence and morphology, and post-surgical change tracking, for a total of 1{,}075{,}092 pairs. Diagnosis tasks emphasize tumour characterization and staging, including BI-RADS and pathology classification, lesion-level signal and kinetic characterization, and associated findings such as lymph-node involvement and tissue invasion, for a total of 680{,}409 pairs. Treatment tasks are grounded in histopathology and treatment-relevant radiology, including prognosis prediction, surgical planning, systemic therapy and biomarker assessment, and WSI-based pathology subtyping, for a total of 100{,}567 pairs. The hierarchical organization ensures that instruction tuning covers the distinct reasoning patterns required across the patient journey. Please check $\S$ \ref{supp:instruction_pipeline} for details.

\noindent\textbf{iii. Textual Description Generation.} The same structured records that feed VQA construction are also reused to synthesise two narrative task types: ground captions and report generation: A \underline{\textit{ground caption}} encodes 2D or 3D spatial coordinates of lesions together with their categories and anatomical locations, providing a multi-dimensional spatial observation for BUS, mammography, CT, and MRI. We construct ground captions by prompting Qwen3-Max to integrate the original clinical report with the available bounding-box annotations into a single visually grounded narrative. \underline{\textit{Medical reports}}, in contrast, condense the multimodal evidence into key findings, tissue characteristics, and diagnostic or therapeutic assessments. After consultation with senior breast oncologists, we organize each generated report into the four ACR-aligned sections \emph{Findings}, \emph{Impression}, \emph{Final Assessment}, and \emph{Management}. The report generation pipeline is iterative: Qwen3-Max first drafts a comprehensive report from the multimodal input, including multiparametric MRI sequences and histopathology WSIs; breast specialists then audit a sample, catalog the recurring reasoning errors and hallucinations, and we refine the modality-specific system prompts before regenerating.

\noindent\textbf{iv. Data Splitting and Verification.} To prevent data leakage while preserving task diversity, we enforce strict patient-level separation between the training set and the evaluation benchmark. We allow the same image to be reused across different task contexts—for instance a BUS image may carry both a BI-RADS label and a lesion-characterization task—but every record derived from a given patient is kept on a single side of the split. The split itself uses stratified sampling on the composite key of modality, task type, and pathology label, so the underlying task and class distributions stay consistent across train and test. Downstream evaluation therefore reflects clinical generalization rather than memorization of shared visual content.

For data reliability we run a multi-stage verification pipeline. Automated heuristic filters first remove hallucinated clinical terms, malformed bounding-box coordinates, instruction-answer conflicts, and near-duplicate QA pairs detected by MinHash similarity above 0.85. Three breast specialists then conduct an independent task-level audit on the held-out test partition (5 random samples per task per specialist), and any task with consistent flags triggers refinement of the generation prompts or filtering rules. \textbf{Notably}, the clinical reliability of all generated data has been audited by clinical experts. In $\S$ \ref{supp:expert_validation}, we report expert agreement on a stratified subset of BreastStage-Bench, including task validity, answer correctness, and clinical consistency. The expert audit covers all modalities and task families, and disagreements are resolved by consensus review.

\section{BreastGPT}
\begin{figure}[!t]
  \centering
  \includegraphics[width=\linewidth]{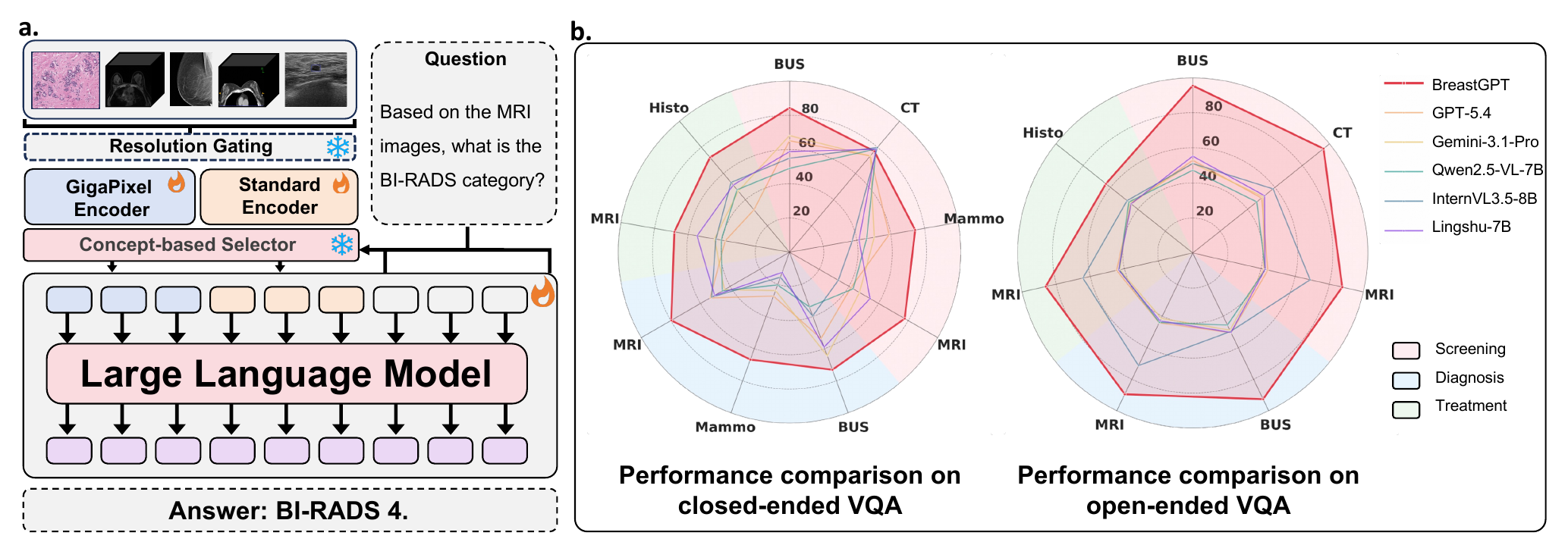}
  \caption{(a) \textbf{Architecture of BreastGPT.} A resolution gating mechanism routes standard radiological images to a ViT-based \textbf{Standard Branch}, and extreme-resolution WSIs to a specialized \textbf{GigaPixel Branch}. The WSI pipeline extracts patch features via a frozen CONCHv1.5 encoder, aggregates global context using LongNet, and compresses the sequence via a universal concept-based token selector before LLM injection. (b) \textbf{Performance Comparison.} BreastGPT demonstrates superior reasoning against state-of-the-art MLLMs across both closed- and open-ended clinical QA tasks.}
  \label{fig:model_architecture}
\end{figure}

BreastGPT is built upon Qwen3-VL~\cite{qwen35vl2025}, a vision-language model with native multi-image support, following the broader Qwen-VL lineage of resolution-aware multimodal modeling~\cite{Qwen-VL,wang2024qwen2,bai2025qwen25vl}. We preserve the LLM backbone architecture and adapt it to breast cancer tasks through stage-aware system prompts, which act as lightweight task routers at inference time. This design allows a single unified model to operate across the screening $\rightarrow$ diagnosis $\rightarrow$ treatment workflow without introducing task-specific heads. Compared with fragmented modality-specific systems, such a unified design shares multimodal representations across tasks while maintaining stage-aware specialization. As a result, BreastGPT supports diverse specialist-style reasoning across 136 task templates within a single instruction-following framework. To handle image scales ranging from standard radiological scans to gigapixel histopathology, we design a dual-branch visual processing architecture modulated by a \textit{modality-aware resolution gating} mechanism. Inputs tagged as WSIs are routed to the GigaPixel branch, while all other radiological modalities (CT, MRI, BUS, mammography) flow through the Standard branch, as illustrated in \cref{fig:model_architecture} a.

\subsection{Dual-Branch Visual Encoding}

\textbf{Standard Encoder for Radiology.} Radiological modalities (CT, MRI, BUS, mammography) are encoded by the native Vision Transformer (ViT) of the Qwen3-VL backbone. This branch is optimized to capture macroscopic anatomical structures, tissue densities, and localized lesion textures that are critical for screening and diagnostic reasoning.

\textbf{GigaPixel Encoder for Pathology.} To integrate extreme-resolution histopathology without overwhelming the LLM context window, we introduce a specialized WSI pathology branch. Prior work on pathology foundation models and WSI transformers has shown that both spatially preserved local morphology and slide-level context are essential for reliable gigapixel image understanding~\cite{anon2025visionb,anon2025visuallanguage,anon2025generalpurpose,anon2025towardsb,anon2025queryaware,anon2025hierarchical,anon2025rethinking,anon2025survey}. Pathology information must therefore be incorporated without collapsing tens of thousands of patch-level features into clinically uninformative representations. We separate WSI processing into two steps before compression:
\begin{itemize}
    \item \textit{Feature Extraction:} WSI tiles at 20$\times$ magnification are encoded by a frozen CONCH v1.5 foundation model~\cite{anon2025visuallanguage}, producing 512-dimensional patch embeddings $\{p_1, \ldots, p_N\}$, where $N$ can reach tens of thousands. Using a frozen encoder retains robust pathology representations from large-scale pretraining while avoiding the computational instability of end-to-end retraining.
    \item \textit{Context Aggregation:} A trainable LongNet dilated attention encoder~\cite{anon2025longnet} aggregates local cellular details and global architectural patterns across the full patch sequence, producing contextualized representations $\{h_1, \ldots, h_N\}$. Because diagnostic cues arise from both local cell morphology and broader tissue organization, LongNet's exponentially growing dilation rates preserve long-range context without the $O(N^2)$ bottleneck of standard attention.
\end{itemize}
Full implementation details (hyperparameters, dilation schedule, projection dimensions) are in $\S$ \ref{supp:gigapixel}.

\subsection{Universal Concept-Based Compression}
A unified model spanning 3D radiological volumes and gigapixel pathology faces a severe token-budget mismatch, as a single WSI patch sequence routinely exceeds tens of thousands of tokens. Rather than resorting to modality-specific pooling or heuristic truncation, we adapt the MMTok multimodal coverage maximization framework~\cite{dong2025mmtok} as a universal concept-based token selector, following recent work on efficient VLM token pruning and compact visual representations~\cite{anon2025adaptive,anon2025learning,anon2025effective,anon2025perlayer,anon2025token}. Given either redundant radiological ViT tokens or contextualized WSI embeddings $\mathbf{X}=\{x_i\}_{i=1}^{N}$, the selector distills the input into exactly $k$ tokens by maximizing a concept-coverage objective:
\begin{equation}
\max_{S \subseteq \{1,\ldots,N\}, |S|=k}
\underbrace{\frac{1}{M}\sum_{j=1}^{M}\max_{i\in S}\mathrm{sim}(t_j,x_i)}_{\text{text--vision coverage}}
+\alpha\underbrace{\frac{1}{N}\sum_{j=1}^{N}\max_{i\in S}\mathrm{sim}(x_j,x_i)}_{\text{vision--vision coverage}},
\end{equation}
where $\{t_j\}_{j=1}^{m}$ are text tokens from the clinical instruction and $\alpha$ balances query relevance with visual representativeness. The first term anchors selection to prompt-relevant clinical concepts, while the second term prevents collapse onto redundant salient regions and encourages coverage of global tissue or anatomical structure. This retains visually distinct, clinically informative patterns, enabling BreastGPT to condense multi-scale visual evidence into a fixed token budget without sacrificing the diversity required for downstream reasoning. The complete calibration and greedy optimization procedure are provided in $\S$ \ref{subsec:mmtok}.

\noindent\textbf{Training.} BreastGPT is trained in two stages: a visual front-end warm-up that freezes the LLM and aligns the ViT and CONCH+LongNet branches with the multimodal token space, followed by end-to-end fine-tuning on all four task formats across every modality. The coverage token selector is training-free and applied identically at training and inference. Detailed optimizer settings, and per-stage trainable/frozen module lists are reported in $\S$ \ref{supp:training_config}.

\begin{table*}[!t]
\caption{VQA performance on BreastStage-Bench. The left half reports \textbf{closed-ended accuracy (\%)} and the right half reports \textbf{open-ended normalized score (\%)} for the same set of models. Open-ended evaluation does not include Mammography. Avg is the mean across the task columns of the corresponding half.}
\label{tab:vqa}
\centering
\scriptsize
\setlength{\tabcolsep}{2pt}
\resizebox{\textwidth}{!}{%
\vspace{-3em}
\begin{tabular}{l c *{9}{c} >{\columncolor{pink!50}}c | *{7}{c} >{\columncolor{pink!50}}c}
\toprule
& & \multicolumn{10}{c|}{\textbf{Closed-ended VQA (Accuracy, \%)}} & \multicolumn{8}{c}{\textbf{Open-ended VQA (normalized Score, \%)}} \\
\cmidrule(lr){3-12}\cmidrule(lr){13-20}
& & \multicolumn{4}{c}{Screening} & \multicolumn{3}{c}{Diagnosis} & \multicolumn{2}{c}{Treatment} & \multicolumn{1}{c}{} & \multicolumn{3}{c}{Screening} & \multicolumn{2}{c}{Diagnosis} & \multicolumn{2}{c}{Treatment} & \multicolumn{1}{c}{} \\
\cmidrule(lr){3-6}\cmidrule(lr){7-9}\cmidrule(lr){10-11}\cmidrule(lr){13-15}\cmidrule(lr){16-17}\cmidrule(lr){18-19}
Model & \#P & BUS & CT & Mam & MRI & BUS & Mam & MRI & MRI & His & \multicolumn{1}{c}{Avg} & BUS & CT & MRI & BUS & MRI & MRI & His & \multicolumn{1}{c}{Avg} \\
\midrule
\hline
\rowcolor{gray!10}
\multicolumn{20}{l}{\textit{Proprietary Models}} \\
\hline
GPT-5.4 & -- & 64.89 & \underline{78.55} & \underline{68.51} & 41.43 & 53.46 & 55.26 & 53.50 & 38.10 & 32.28 & 54.00 & 53.43 & 49.34 & 59.85 & 50.42 & 59.20 & 58.07 & 44.73 & 53.58 \\
Claude-opus-4-6 & -- & 50.21 & 72.00 & 39.57 & 50.48 & 38.83 & 7.66 & 45.10 & 41.27 & 25.94 & 41.23 & 43.32 & 42.77 & 43.75 & 42.80 & 44.42 & 42.91 & 40.84 & 42.97 \\
Claude-sonnet-4-6 & -- & 65.53 & 68.00 & 37.02 & 40.00 & 54.79 & 19.67 & 55.94 & 42.86 & 36.30 & 46.68 & 43.63 & 43.12 & 44.31 & 43.00 & 44.98 & 42.62 & 41.06 & 43.25 \\
Gemini-3.1-Flash & -- & 55.11 & 42.67 & 49.68 & 38.57 & 37.60 & 24.32 & 31.82 & 48.41 & 35.81 & 40.44 & 48.76 & 48.21 & 47.38 & 45.80 & 47.53 & 44.15 & 42.81 & 46.38 \\
Gemini-3.1-Pro & -- & 68.09 & 73.33 & 50.21 & 47.14 & 64.36 & 23.87 & 43.88 & 44.44 & 46.53 & 51.32 & 51.19 & 51.46 & 41.60 & 48.94 & 41.39 & 42.80 & 45.77 & 46.16 \\
\hline
\rowcolor{gray!10}
\multicolumn{20}{l}{\textit{Open-Source Models}} \\
\hline
Qwen2.5-VL & 3B & 46.81 & 51.39 & 33.51 & 45.71 & 30.85 & 12.76 & 51.75 & 44.44 & 51.52 & 40.97 & 49.52 & 48.39 & 49.36 & 47.36 & 50.54 & 47.04 & 44.80 & 48.14 \\
Qwen2.5-VL & 7B & 49.15 & \textbf{79.27} & 44.47 & 44.76 & 34.31 & 14.41 & 46.85 & 37.30 & 47.62 & 44.24 & 46.20 & 46.86 & 49.47 & 44.99 & 49.67 & 44.70 & 43.98 & 46.55 \\
Qwen3-VL & 4B & 52.13 & 75.39 & 41.91 & 49.05 & 35.37 & 21.17 & 48.08 & 43.65 & 38.61 & 45.04 & 45.20 & 55.49 & 47.71 & 44.93 & 51.46 & 48.98 & 41.26 & 47.86 \\
Qwen3-VL & 8B & 57.87 & \underline{78.55} & 39.68 & 51.43 & 48.14 & 25.83 & 47.90 & 34.92 & 47.02 & 47.93 & 44.94 & 48.53 & 44.85 & 44.34 & 46.24 & 44.15 & 41.21 & 44.89 \\
MiMo-VL & 7B & 15.11 & 23.52 & 8.62 & 6.67 & 14.89 & 6.16 & 0.35 & 0.79 & 19.98 & 10.68 & 42.49 & 55.98 & 63.70 & 41.69 & 65.66 & 63.08 & 39.86 & 53.21 \\
InternVL3.5 & 8B & 51.91 & 77.70 & 35.85 & 34.76 & 42.55 & 16.22 & 52.27 & 44.44 & 52.98 & 45.41 & 50.74 & 58.05 & 56.56 & 48.80 & 58.89 & 55.94 & 46.50 & 53.64 \\
\hline
\rowcolor{gray!10}
\multicolumn{20}{l}{\textit{Medical-Specific Models}} \\
\hline
Lingshu & 7B & 58.94 & \underline{78.55} & 39.89 & 54.29 & 58.24 & 8.56 & 45.28 & \underline{58.73} & 51.52 & 50.44 & 52.60 & 49.84 & 53.41 & 48.95 & 53.76 & 49.60 & 43.69 & 50.26 \\
HuatuoGPT-V & 7B & 45.74 & 71.39 & 43.09 & 39.05 & 35.11 & 9.61 & 47.73 & 45.24 & 49.09 & 42.89 & 51.86 & 51.47 & 55.68 & 49.20 & 55.08 & 52.93 & 45.78 & 51.71 \\
\rowcolor{cyan!15}\textbf{Qwen3-VL (SFT)} & \textbf{8B} & 72.34 & 76.73 & 65.74 & \underline{78.10} & 71.01 & \underline{59.76} & \underline{75.87} & 53.97 & 60.41 & \cellcolor{pink!50}68.21 & 94.43 & \textbf{95.38} & \underline{94.57} & 91.70 & \underline{95.28} & \underline{88.36} & 57.96 & \cellcolor{pink!50}\underline{88.24} \\
\rowcolor{cyan!15}\textbf{BreastGPT (cluster)} & \textbf{8B} & \textbf{86.81} & 77.21 & \textbf{75.00} & \textbf{82.86} & \underline{77.13} & \textbf{68.32} & \textbf{81.12} & \textbf{61.11} & \textbf{71.38} & \cellcolor{pink!50}\textbf{75.66} & \textbf{95.97} & \underline{95.29} & \textbf{95.48} & \underline{93.24} & \textbf{95.72} & \textbf{89.93} & \textbf{63.80} & \cellcolor{pink!50}\textbf{89.92} \\
\rowcolor{cyan!15}\textbf{BreastGPT (learn)} & \textbf{8B} & \underline{84.47} & 71.03 & \underline{68.51} & 75.71 & \textbf{78.46} & 55.26 & 73.95 & 57.14 & \underline{71.25} & \cellcolor{pink!50}\underline{70.64} & \underline{95.86} & 94.73 & 87.12 & \textbf{93.57} & 85.22 & 81.78 & \underline{63.36} & \cellcolor{pink!50}85.95 \\
\bottomrule
\end{tabular}%
}
\vspace{1mm}
\begin{minipage}{\linewidth}
\footnotesize
\textit{Note:} BUS = Breast Ultrasound, CT = Computed Tomography, Mam = Mammography, His = Histopathology.  Mammography is omitted from the right half because it has no open-ended evaluation set.
\end{minipage}
\end{table*}
\section{Experiments}

\subsection{Experimental Setups}
\noindent\textbf{Baselines.} We compare BreastGPT with three groups of representative vision--language models on BreastStage-Bench: \textit{1) proprietary frontier models}: GPT-5.4, Claude-opus-4-6, Claude-sonnet-4-6, Gemini-3.1-Flash, and Gemini-3.1-Pro, queried through their official APIs under identical stage-specific system prompts; these models represent frontier model families commonly used in recent medical MLLM benchmarks~\cite{hurst2024gpt,team2023gemini}; \textit{2) open-source general-purpose VLMs} at comparable scale to BreastGPT: Qwen2.5-VL-Instruct (3B/7B)~\cite{bai2025qwen25vl}, Qwen3-VL-Instruct (4B/8B)~\cite{qwen35vl2025}, MiMo-VL-SFT~\cite{coreteam2025mimovltechnicalreport}, and InternVL3.5; \textit{3) medical-specific VLMs}: Lingshu~\cite{anon2025generalist} and HuatuoGPT-V~\cite{chen2024huatuogpt}, with additional context from prior biomedical and medical VLMs~\cite{li2023llavamed,he2024meddr,lin2025healthgpt}. All baselines are evaluated zero-shot on the same instruction-formatted prompts used for BreastGPT, ensuring that differences in performance reflect model capability rather than prompt engineering. Detailed per-baseline configuration is listed in $\S$ \ref{supp:baseline_config}.

\noindent\textbf{BreastGPT variants.}
\label{sec:experiment}
We report two variants that differ only in how the final 128 tokens are obtained to verify the effect of our training-free selection strategy: BreastGPT (cluster) uses the training-free greedy coverage selector, while BreastGPT (learn) replaces it with a learnable cross-attention selector inspired by the Perceiver Resampler in Flamingo~\cite{alayrac2022flamingo}. Both variants share the same backbone, branches, and training data.

\subsection{Evaluation Metrics}
We employ a comprehensive suite of metrics tailored to the diverse task formats within BreastStage-Bench. For closed-ended tasks, we report standard classification accuracy (ACC). For open-ended reasoning and report generation tasks, we report BERTScore F1, BLEU, and ROUGE-1, alongside a weighted aggregate score (0.5 BERTScore F1, 0.25 BLEU, and 0.25 ROUGE-1). We assign a predominant weight to BERTScore F1 because it captures deep semantic equivalence rather than surface-level lexical overlap, a critical requirement for evaluating clinically constrained free-text responses. For visual grounding tasks, we report mean Intersection over Union (IoU); a sample without a ground-truth bounding box is credited as IoU\,$=$\,1 when the model also abstains from emitting a box (TN) and 0 when it hallucinates one (FP). Please check $\S$ \ref{supp:eval_metrics} for details.

\subsection{Evaluation Results}

\noindent\textbf{Overall observations.}
Across the main results in \cref{tab:vqa,tab:caption_report}, BreastGPT achieves the strongest performance across clinical stages and task formats, with particularly large gains over both general-purpose and medical-specific VLMs.
On closed-ended VQA, BreastGPT (cluster) reaches 75.66\% average accuracy, exceeding the strongest proprietary baseline, GPT-5.4, by 21.66 points.
On open-ended VQA, it reaches 89.92\%, remaining more than 25 points ahead of all models in the extended comparison.
For generation tasks, BreastGPT improves BUS captioning from 51.48 to 79.32 weighted score and MRI report generation from 55.16 to 67.67.
On 3D grounding for CT and MRI, BreastGPT is the only model that produces non-trivial volumetric bounding boxes.
Full comparisons with the extended baselines are provided in $\S$ \cref{tab:vqa_extended,tab:caption_report_extended}, and per-modality 3D grounding IoU is reported in $\S$ \cref{tab:caption_3d_iou}.

\begin{table*}[t]
\caption{Caption and Report Generation Performance (\%). IoU = mean grounding IoU on the grounded captioning task, ``-'' indicates the model produces no usable bbox in that modality. Wtd = weighted composite ($0.5\times$BERT $+$ $0.25\times$BLEU $+$ $0.25\times$R-1). IoU on the 3D modalities (CT, MRI) is reported in \cref{tab:caption_3d_iou}.}
\label{tab:caption_report}
\centering
\small
\resizebox{\textwidth}{!}{%
\vspace{-3em}
\begin{tabular}{p{3.5cm} c
  *{2}{>{\centering\arraybackslash}p{1.7cm}}
  >{\centering\arraybackslash}p{1.7cm}
  *{2}{>{\centering\arraybackslash}p{1.7cm}}
  >{\centering\arraybackslash}p{1.7cm}
  >{\centering\arraybackslash}p{1.7cm}|
  >{\centering\arraybackslash}p{1.7cm}}
\toprule
& & \multicolumn{7}{c|}{\textbf{Caption}} & \textbf{Report} \\
\cmidrule(lr){3-9}\cmidrule(lr){10-10}
& & \multicolumn{2}{c}{\textbf{BUS}} & \textbf{CT} & \multicolumn{2}{c}{\textbf{Mammo}} & \textbf{Histo} & \textbf{MRI} & \textbf{MRI} \\
\cmidrule(lr){3-4}\cmidrule(lr){5-5}\cmidrule(lr){6-7}\cmidrule(lr){8-8}\cmidrule(lr){9-9}\cmidrule(lr){10-10}
Model & \#P & IoU & Wtd & Wtd & IoU & Wtd & Wtd & Wtd & Wtd \\
\midrule
\hline
\rowcolor{gray!10}
\multicolumn{10}{l}{\textit{Proprietary Models}} \\
\hline
GPT-5.4 & -- & 6.68 & 47.66 & 43.95 & 9.08 & 45.10 & 44.36 & 47.00 & 55.16 \\
Claude-opus-4-6 & -- & 11.41 & 46.99 & 43.68 & \underline{12.06} & 47.04 & 44.94 & 47.75 & 48.56 \\
Claude-sonnet-4-6 & -- & 7.40 & 46.60 & 42.82 & -- & 44.79 & 44.21 & 49.30 & 49.30 \\
Gemini-3.1-Flash & -- & \underline{56.20} & 51.48 & 46.01 & 7.52 & 49.29 & 44.89 & 51.45 & 51.03 \\
Gemini-3.1-Pro & -- & -- & 45.60 & 44.64 & -- & 42.36 & 45.84 & 54.86 & 54.86 \\
\hline
\rowcolor{gray!10}
\multicolumn{10}{l}{\textit{Open-Source Models}} \\
\hline
Qwen2.5-VL & 3B & 2.67 & 46.84 & 46.88 & 1.38 & 48.36 & 45.71 & 50.69 & 49.59 \\
Qwen2.5-VL & 7B & 6.38 & 46.79 & 47.31 & 1.47 & 48.11 & 45.07 & 51.95 & 50.71 \\
Qwen3-VL & 4B & 33.64 & 46.68 & 41.66 & 4.29 & 46.19 & 45.38 & 37.66 & 49.46 \\
Qwen3-VL & 8B & 29.35 & 46.97 & 47.91 & 4.06 & 47.09 & 46.03 & 49.57 & 50.49 \\
MiMo-VL & 7B & 12.35 & 43.22 & 46.14 & 2.22 & 43.80 & 43.98 & 47.32 & 46.55 \\
InternVL3.5 & 8B & 0.35 & 46.97 & 47.69 & 0.17 & 45.72 & 45.77 & 49.28 & 49.12 \\
\hline
\rowcolor{gray!10}
\multicolumn{10}{l}{\textit{Medical-Specific Models}} \\
\hline
Lingshu & 7B & 5.00 & 48.43 & 49.45 & 0.55 & 49.14 & 46.86 & 51.84 & 51.14 \\
HuatuoGPT-V & 7B & 0.21 & 47.46 & 48.92 & 0.47 & 48.84 & 46.24 & 51.46 & 51.00 \\
\rowcolor{cyan!15}
\textbf{Qwen3-VL (SFT)} & \textbf{8B} & 72.79 & 76.61 & 72.72 & 8.62 & \underline{76.68} & 62.42 & \underline{65.21} & \underline{66.10} \\
\rowcolor{cyan!15}
\textbf{BreastGPT (cluster)} & \textbf{8B} & \textbf{79.59} & \textbf{79.32} & \underline{73.16} & \textbf{23.14} & \textbf{77.64} & \underline{66.78} & \textbf{67.69} & \textbf{67.67} \\
\rowcolor{cyan!15}
\textbf{BreastGPT (learn)} & \textbf{8B} & -- & \underline{79.16} & \textbf{77.07} & -- & 76.54 & \textbf{68.11} & -- & 66.04 \\
\bottomrule
\end{tabular}%
}
\end{table*}

\noindent\textbf{Proprietary frontier models underperform on breast workflows.}
Despite their general multimodal capabilities, proprietary frontier models remain substantially below BreastGPT on breast-cancer workflow tasks.
Even the strongest proprietary baseline, GPT-5.4, reaches only 54.00\% average accuracy on closed-ended VQA, more than 20 points behind BreastGPT.
The gap is especially pronounced on mammography-based diagnosis, where Claude-opus-4-6, Claude-sonnet-4-6, and Gemini-3.1-Pro all perform poorly.
These results suggest that breast workflows require domain-specific visual and clinical knowledge, including BI-RADS terminology, molecular subtyping conventions, and multiparametric MRI interpretation, that is not reliably captured by general web-scale multimodal pretraining.

\noindent\textbf{Medical-specific VLMs offer no clear advantage.} Lingshu and HuatuoGPT-V perform comparably to general-purpose 7--8B VLMs on closed-ended VQA (50.44\% and 42.89\%), and sometimes underperform them. This indicates that existing medical pretraining corpora do not transfer specifically to breast cancer workflow tasks, motivating the need for a workflow-aligned dataset like BreastStage.

\noindent\textbf{Cluster vs. learn variant.} BreastGPT (cluster), which uses the training-free greedy coverage selector, edges out BreastGPT (learn) on both closed-ended VQA (75.66\% vs. 70.64\%) and open-ended VQA (89.92\% vs. 85.95\%), despite having no additional learnable parameters for token selection. This confirms that the submodular coverage objective is a strong inductive bias for WSI and multimodal token compression, and that a learnable retriever does not trivially outperform a principled, training-free procedure in the breast workflow setting.

\subsection{Ablation Studies}
We conduct three ablation studies to verify the contribution of individual components.

\noindent\textbf{Necessity of the GigaPixel branch.}
To isolate the contribution of the GigaPixel branch, we compare BreastGPT with a controlled Qwen3-VL-8B SFT baseline that uses the same training data and recipe but processes WSIs through the standard ViT branch.
Specifically, each slide is tiled into non-overlapping $512{\times}512$ patches, from which 32 patches are randomly sampled during both training and inference.
This baseline controls for the backbone and supervision while removing the dedicated gigapixel representation and selection mechanism. As reported in \cref{tab:vqa,tab:caption_report}, the gap between this baseline and BreastGPT is modest on radiology modalities, where the source images already fit naturally into a ViT-style pipeline.
On histopathology, however, the gap becomes substantial: the 32-patch baseline reaches 60.41\% closed-ended accuracy and 62.42 captioning weighted score, compared with 71.38\% and 66.78 for BreastGPT (cluster).
Because 32 random $512{\times}512$ tiles cover only a tiny fraction of a typical breast WSI, slide-level cues related to subtype, grade, and treatment response can be easily missed.
The CONCHv1.5{+}LongNet branch and coverage selector recover this information by reasoning over a globally contextualized representation of the whole slide rather than a sparse random sample.
A complementary per-component ablation inside the GigaPixel branch is provided in $\S$ Table~\ref{tab:wsi_ablation_numerical}.

\begin{figure}[!t]
  \centering
  \includegraphics[width=\linewidth]{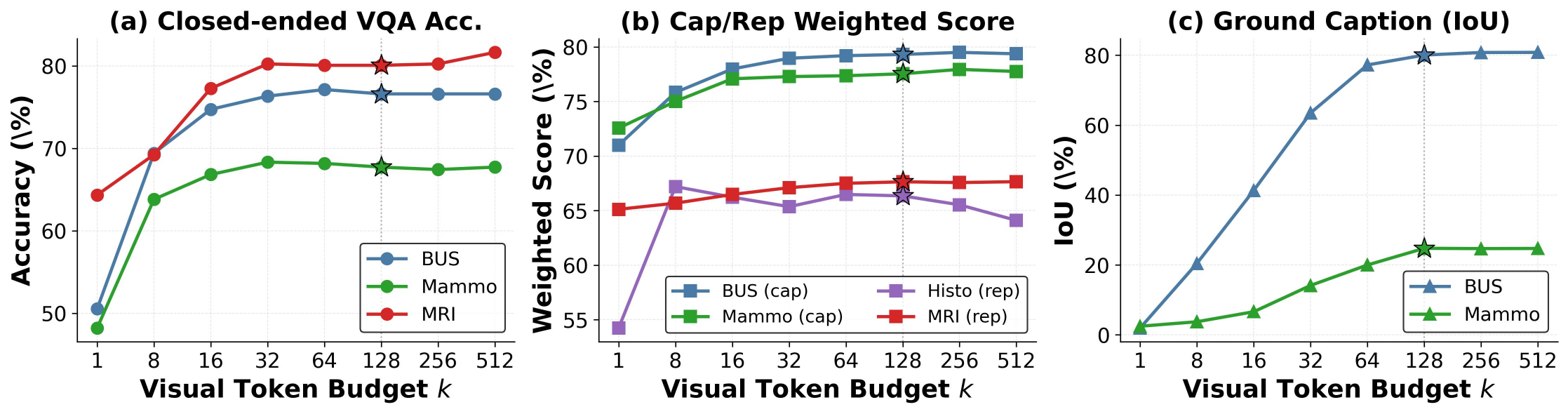}
  \caption{Visual token budget sweep on BreastStage-Bench. \textbf{(a)} Closed-ended VQA accuracy on diagnosis-stage tasks for BUS, mammography, and MRI. \textbf{(b)} Caption and report weighted score for BUS, mammography, histopathology report, and MRI report. \textbf{(c)} Ground caption IoU on BUS and mammography. Dotted vertical lines and stars mark the chosen operating point $k=128$.}
  \label{fig:token_budget}
\end{figure}
\begin{wrapfigure}{r}{0.4\linewidth}
  \vspace{-1.2em}
  \centering
  \includegraphics[width=\linewidth]{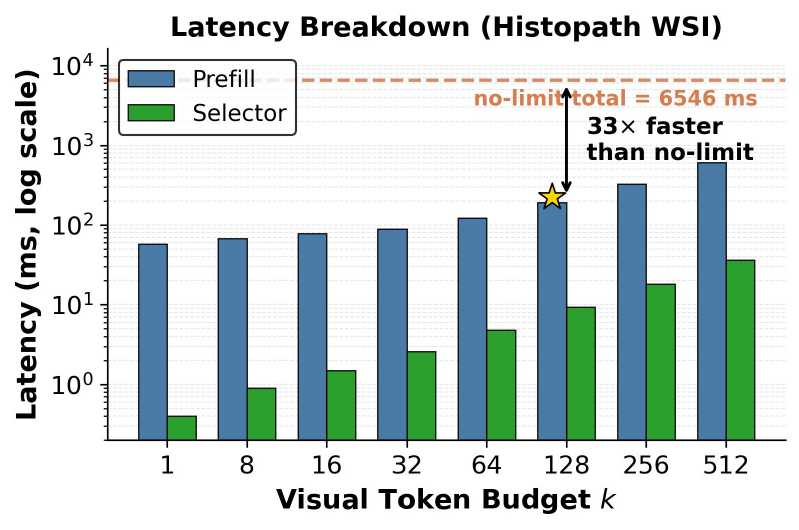}
  \caption{GPU latency on a histopathology WSI. Two bars per $k$ give prefill and selector latency. The dashed line is the no-limit baseline that bypasses the selector; its gap to the chosen-$k$ prefill bar is the latency saved by selection. Log-scaled $y$-axis.}
  \label{fig:latency}
  \vspace{-1.2em}
\end{wrapfigure}
\noindent\textbf{Visual token budget sensitivity.} We vary the visual token budget from $k{=}1$ to $k{=}512$ across five modalities and three task families in BreastStage-Bench, with full results reported in $\S$ \cref{tab:token_budget_vqa_full,tab:token_budget_caption_full}. \textit{Closed-ended VQA.} As shown in \cref{fig:token_budget}a, diagnosis-stage VQA saturates early, with most modalities reaching a plateau by $k{=}32$--$64$. \textit{Caption and report generation.} In \cref{fig:token_budget}b, performance peaks around $k{=}128$. Increasing the budget beyond this point brings negligible gains for radiology modalities and can even degrade histopathology performance, suggesting that excessive WSI tokens may introduce redundant or distracting information. \textit{Grounded captioning.} The grounding task is more token-demanding (\cref{fig:token_budget}c). A single visual token is insufficient to capture the spatial extent of a bounding region, but IoU improves rapidly with larger budgets and recovers over 99\% of the $k{=}512$ performance by $k{=}128$.


\textbf{Takeaway.} We therefore set $k{=}128$ as the unified budget: it lies on the saturation plateau of VQA and generation, retains nearly all of the $k{=}512$ grounded-captioning IoU, and keeps radiology and pathology inputs at the same downstream cost.

\noindent\textbf{GPU time.} \cref{fig:latency} measures inference latency on a representative histopathology WSI with 5{,}987 patches. At the chosen budget $k{=}128$, LLM prefill takes 191.4 ms and the selector adds only 9.3 ms, yielding approximately 200.6 ms total latency. By contrast, feeding all 5{,}987 patch tokens directly to the LLM increases latency to 6.5 s, or about $33\times$ slower than the selected-token setting. Increasing the budget to $k{=}512$ raises prefill latency to 605.7 ms while improving task quality by less than $1\%$ over $k{=}128$. Thus, the selector adds a small overhead but removes the much larger LLM cost that would otherwise dominate WSI inference. Full per-$k$ numbers are provided in $\S$ \cref{tab:inference_efficiency}.

\noindent\textbf{More results.} Additional per-modality results and inference-efficiency analysis are provided in $\S$ \ref{supp:additional_results}. More detailed ablations of the WSI branch and visual token budget are reported in $\S$ \ref{supp:detailed_ablation}. Qualitative examples and case studies are presented in $\S$ \ref{supp:qualitative}.

\section{Discussion and Conclusion}
\noindent
Breast cancer care is, in essence, a tightly coupled clinical workflow spanning screening, diagnosis, and treatment, yet existing medical multimodal LLMs are largely optimized for individual stages and struggle to support the cross-stage reasoning that real clinical practice demands. To close this gap, we tackle the problem from both the data and the model side: we propose \textbf{BreastStage}, a large-scale multimodal instruction dataset and benchmark organized around the screening--diagnosis--treatment pathway, and \textbf{BreastGPT}, a unified vision-language model that pairs stage-aware role prompting with a dual-branch visual encoder, enabling cross-scale modeling from standard radiology to gigapixel pathology images.
On BreastStage-Bench, both general-purpose and medical-specific MLLMs fall short on cross-stage reasoning, while our workflow-aligned design substantially closes this gap across modalities and task formats. We hope these resources help shift medical multimodal modeling from isolated tasks toward clinical-workflow-centred reasoning.

\textbf{Limitation and Future Work.} 
While BreastStage covers screening, diagnosis, and treatment planning, most training samples are collected from different patient cohorts rather than from longitudinal records of the same patients. Therefore, the current training corpus supports workflow-aligned supervision at the task and modality level, but does not fully capture patient-level temporal continuity across the entire care pathway. Notably, a subset in BreastStage-bench includes full-workflow cases from the same patients, allowing us to partially examine whether the model can integrate clinically connected evidence across stages. Building a truly per-patient cross-stage corpus is a field-level bottleneck, requiring linked records across departments, multi-year follow-up windows, and IRB approval for identifiable longitudinal data and we are working for this.

\newpage

\bibliographystyle{assets/plainnat}
\bibliography{paper}

\newpage
\beginappendix
\lstdefinelanguage{jsonsch}{
  basicstyle=\scriptsize\ttfamily,
  breaklines=true,
  columns=fullflexible,
  showstringspaces=false,
  keywordstyle=\color{blue!55!black},
  stringstyle=\color{red!55!black},
  commentstyle=\color{green!45!black}\itshape,
  morecomment=[l]{//},
  morestring=[b]",
  xleftmargin=2pt, xrightmargin=2pt,
  aboveskip=2pt, belowskip=2pt}
\newtcolorbox{promptbox}[2][]{%
  enhanced, colback=#2!8, colframe=#2!55!black, sharp corners=northwest,
  fonttitle=\bfseries\footnotesize, title=#1,
  coltitle=white, colbacktitle=#2!55!black,
  fontupper=\scriptsize, boxrule=0.5pt, arc=2pt,
  left=4pt, right=4pt, top=3pt, bottom=3pt,
  attach boxed title to top left={xshift=4pt, yshift=-1pt},
  boxed title style={sharp corners, boxrule=0pt},
  before upper={\setlength{\parskip}{2pt}\setlength{\parindent}{0pt}}}
\newtcolorbox{personabox}[2][]{%
  enhanced, colback=#2!8, colframe=#2!55!black, sharp corners=northwest,
  fonttitle=\bfseries\footnotesize, title=#1,
  coltitle=white, colbacktitle=#2!55!black,
  fontupper=\scriptsize, boxrule=0.5pt, arc=2pt,
  left=4pt, right=4pt, top=3pt, bottom=3pt,
  attach boxed title to top left={xshift=4pt, yshift=-1pt},
  boxed title style={sharp corners, boxrule=0pt},
  before upper={\setlength{\parskip}{2pt}\setlength{\parindent}{0pt}},
  equal height group=personarow}
\newtcolorbox{promptboxA}[2][]{%
  enhanced, colback=#2!8, colframe=#2!55!black, sharp corners=northwest,
  fonttitle=\bfseries\footnotesize, title=#1,
  coltitle=white, colbacktitle=#2!55!black,
  fontupper=\scriptsize, boxrule=0.5pt, arc=2pt,
  left=4pt, right=4pt, top=3pt, bottom=3pt,
  attach boxed title to top left={xshift=4pt, yshift=-1pt},
  boxed title style={sharp corners, boxrule=0pt},
  before upper={\setlength{\parskip}{2pt}\setlength{\parindent}{0pt}},
  equal height group=qcrow1}
\newtcolorbox{promptboxB}[2][]{%
  enhanced, colback=#2!8, colframe=#2!55!black, sharp corners=northwest,
  fonttitle=\bfseries\footnotesize, title=#1,
  coltitle=white, colbacktitle=#2!55!black,
  fontupper=\scriptsize, boxrule=0.5pt, arc=2pt,
  left=4pt, right=4pt, top=3pt, bottom=3pt,
  attach boxed title to top left={xshift=4pt, yshift=-1pt},
  boxed title style={sharp corners, boxrule=0pt},
  before upper={\setlength{\parskip}{2pt}\setlength{\parindent}{0pt}},
  equal height group=qcrow2}
  
{\centering \Large \textbf{BreastGPT: A Multimodal Large Language Model
} \par}
\vspace{-0.2cm}
{\centering \Large \textbf{for the Full Spectrum of Breast Cancer Clinical Routine} \par}

\vspace{-0.3cm}

\addtocontents{toc}{\protect\setcounter{tocdepth}{2}}
{\renewcommand{\baselinestretch}{1.7}\selectfont
\tableofcontents
\par}

\section{Supplementary Overview}\label{supp:overview}

This supplementary material provides the technical and procedural details that support the main claims of the paper. We first describe the construction of \textbf{BreastStage}, including data provenance, modality coverage, instruction generation, quality control, and task taxonomy. We then provide implementation details for \textbf{BreastGPT}, with particular emphasis on the GigaPixel WSI branch and the coverage-maximizing visual token selector that enables a unified token budget across standard radiology images and gigapixel pathology slides. Finally, we describe the BreastStage-Bench evaluation protocol, additional quantitative analyses, qualitative examples, limitations, ethics, and release plans.

This supplementary is intended to make three aspects of the paper auditable: (i) how heterogeneous breast imaging data are mapped into a stage-aware clinical workflow, (ii) how BreastGPT processes image scales that differ by several orders of magnitude, and (iii) how benchmark results, ablations, and error analyses are produced.

\subsection{LLM Usage Statement}\label{supp:llm_usage}

In this work, we primarily employ LLMs in two aspects: (i) data construction, where Qwen2.5-VL-72B and Qwen3-Max drive the orchestration pipeline described in \cref{supp:instruction_pipeline}; and (ii) manuscript polishing, where we use LLMs to improve grammar and clarity of the writing without altering the technical claims.

\subsection{Ethics Approval and Privacy}\label{supp:ethics_curation}

BreastStage combines two kinds of sources with different ethical pathways. The BUS, mammography, CT, and histopathology subsets come entirely from public datasets that were de-identified by their original publishers and released under their respective data-use terms; we use them within those terms and do not perform any additional re-identification of individual samples. The MRI subset is the only institutional cohort: it was acquired from two collaborating hospitals under IRB approval, with patient consent obtained where required by the contributing institutions.

\paragraph{De-identification of the institutional MRI cohort.}
Because the MRI cohort is the only data source that originates inside hospitals, our de-identification protocol is scoped specifically to it. Before MRI data leave either source institution, identifying information is removed at three levels:
\begin{itemize}
    \item \textbf{DICOM header level.} Patient names, dates of birth, medical-record numbers, hospital identifiers, referring-physician fields, and exam dates are stripped; absolute timestamps are converted to relative offsets so longitudinal ordering is preserved without disclosing calendar dates.
    \item \textbf{Image content level.} The MRI volumes are reconstructed within a breast-coil field of view that does not include the patient's face, so no additional image-side scrubbing is required.
    \item \textbf{Free-text level.} The accompanying Chinese radiology and pathology reports are passed through the bilingual parsers in Figure~\ref{fig:prompt_templates}, which emit a strictly schema-bounded English JSON so that no free-text PHI (referring physician name, hospital department, etc.)\ propagates into the released instruction-following pairs.
\end{itemize}

\paragraph{Access control.}
Access to identifiable raw MRI data is restricted to authorised research personnel under signed data-use agreements. The public mirror of BreastStage will release only the de-identified, instruction-formatted records and the derived bounding-box / mask annotations; raw DICOMs from the institutional MRI cohort are not redistributed. The other four subsets (BUS, mammography, CT, histopathology) are accessed via their original public-dataset distribution channels.

\paragraph{Risks acknowledged.}
Because the model is trained on clinical imaging data, performance may vary across institutions, scanners, demographic groups, and disease subtypes; we recommend that any downstream use perform site-specific validation before clinical deployment. The benchmark is intended to encourage transparent comparison of breast cancer workflow models, not to replace clinician judgement.

\subsection{Additional Limitations}\label{supp:additional_limitations}

\paragraph{Dataset.}
While BreastStage covers five major imaging modalities routinely used in breast cancer screening and diagnosis, it does not currently include molecular imaging modalities such as PET/CT, which are often used for staging, metastasis assessment, and treatment response evaluation in advanced disease. Emerging modalities including photoacoustic imaging are also beyond the current scope. In addition, the present workflow primarily focuses on imaging-centered decision support, while longitudinal clinical information such as treatment trajectories, genomic assays, laboratory tests, and medication records remains only partially available across cohorts. Future work should extend BreastStage toward longitudinal multimodal patient records and investigate whether workflow-aligned MLLMs can support temporal clinical reasoning for treatment response evaluation, recurrence monitoring, and survivorship management.

\paragraph{Clinical deployment.}
BreastGPT is currently developed as a research prototype and has not been clinically validated or reviewed by regulatory authorities. Any use in real-world clinical workflows would require prospective clinical evaluation, compliance with applicable regulatory and data-governance requirements, integration with existing hospital information systems, and continuous post-deployment surveillance to monitor model drift, performance degradation, and potential safety risks. Accordingly, BreastGPT should be regarded as a research and benchmarking foundation model, not as an autonomous diagnostic or treatment-planning system.

\subsection{Compute Resources}\label{supp:compute}

\paragraph{Model training.}
BreastGPT training was carried out on 32 NVIDIA H100 GPUs (4 nodes $\times$ 8 GPUs) with DeepSpeed ZeRO-2, FlashAttention, and bfloat16 precision. End-to-end wall-clock training time across the two stages was 3 days, 7 hours, 19 minutes, 53 seconds, corresponding to approximately 2{,}530 H100-hours.

\paragraph{Dataset construction and evaluation.}
Visual quality control with Qwen2.5-VL-72B and Qwen3-Max-driven instruction generation are run through external APIs and shared infrastructure; per-call API expenditure was not separately tracked. Specialist annotation of MRI tumours (10 board-certified breast radiologists) and the 3-specialist Bench audit were conducted by clinical collaborators rather than billed compute.

\subsection{Code and Data Availability}\label{supp:code_data}

\paragraph{Code.}
A snapshot of the BreastGPT codebase, including training scripts, evaluation pipelines, and BreastStage data-construction tools, is available at \url{https://yangyy-liu.github.io/BreastGPT.io}. Pretrained BreastGPT model weights will be released on ModelScope at \url{https://www.modelscope.cn/models/YYangYang/BreastGPT-8B}.

\paragraph{Data.}
The BreastStage dataset and BreastStage-Bench evaluation suite will be made publicly available at \url{https://www.modelscope.cn/datasets/YYangYang/BreastStage}. For components derived from institutional clinical data, release will follow the corresponding IRB protocol, de-identification requirements, and data-use agreements. When raw image release is restricted, we will provide metadata, task definitions, evaluation code, and access instructions.

\section{BreastStage Curation Details}\label{supp:dataset_curation}

\subsection{Data Sources and Modality Coverage}\label{supp:data_sources}

As described in Section 2 of the main paper, BreastStage integrates 17 sub-datasets across 5 imaging modalities. The goal of this curation is not merely to aggregate public resources, but to reorganize heterogeneous imaging evidence around the clinical progression of breast cancer care: screening, diagnosis, and treatment. This section details the source datasets, selection criteria, quality control procedures, and annotation protocols used to construct each modality subset.

\paragraph{CT Imaging.}
Computed tomography volumes are sourced from the CT-RATE dataset~\cite{ctrate2024}, which contains 25,692 non-contrast chest CT volumes from 21,304 unique patients. Since breast cancer predominantly affects women, we first filter the dataset to retain only female patients (sex metadata field equal to ``F''). We then deploy Qwen2.5-VL-72B~\cite{bai2025qwen25vl} as an automated quality inspector, assigning confidence scores (0--10 scale) based on image quality, anatomical coverage, and the presence of breast tissue in the field of view; volumes scoring below the threshold are excluded. For breast cancer screening and diagnosis, we further pass each volume through DRT-M3D~\cite{zhou2026dualres} to generate automated tumor segmentations and risk assessments, identifying suspicious breast lesions, calcifications, and lymph node involvement visible in the chest CT field of view. After quality filtering and clinical relevance screening, the CT subset of BreastStage contains 20{,}546 reconstructed volumes from female patients (Table~\ref{tab:source_summary}), covering both screening-stage opportunistic detection and diagnosis-stage tumor characterization tasks.

\paragraph{Breast Ultrasound (BUS).}
BUS images are sourced from the \textbf{BUS-CoT} dataset~\cite{buscot2025}, the first chain-of-thought reasoning breast ultrasound dataset that covers \emph{all} 99 histopathology categories defined by the WHO classification of breast tumors. BUS-CoT contains 11,439 images of 10,019 lesions from 4,838 patients across 18 different ultrasound device types, including B-mode US, Doppler US, and Elastography. Images are collected from open-access publications, publicly available case studies (Radiopaedia, PubMed), and open-access biopsy-confirmed repositories, with each case annotated through a rigorous five-level protocol by six senior breast ultrasound radiologists with 8--26 years of clinical experience: (1) \emph{Observation} (lesion presence, calcification presence), (2) \emph{Feature} (boundary, edge, echo characteristics, calcification feature), (3) \emph{Diagnosis} (BI-RADS score), (4) \emph{Pathology} (benign/malignant + WHO histopathology subtype), and (5) \emph{Chain-of-Thought reasoning} explicitly linking imaging features to the final pathological label. Table~\ref{tab:bus_cot_pathology} summarizes the dataset's pathology distribution.

\begin{table}[H]
\centering
\caption{Pathology distribution of the BUS-CoT source dataset~\cite{buscot2025}, used as the BUS data source for BreastStage. Numbers in parentheses indicate per-subtype lesion counts.}
\label{tab:bus_cot_pathology}
\small
\begin{tabular}{l p{4.5cm} p{4.5cm} l}
\toprule
& \textbf{Benign (4{,}856)} & \textbf{Malignant (4{,}814)} & \textbf{Others (349)} \\
\midrule
\multirow{6}{*}{\rotatebox{90}{\textit{Categories}}}
& Fibroadenoma (1{,}047) & Invasive ductal carcinoma (896) & Others (349) \\
& Phyllodes tumour (74) & Invasive lobular carcinoma (75) & \\
& Intraductal papilloma (72) & Ductal carcinoma in situ (72) & \\
& Atypical ductal hyperplasia (62) & Mucinous carcinoma (64) & \\
& Radial scar (56) & Paget disease of the breast (57) & \\
& Other benign (3{,}545) & Other malignant (3{,}650) & \\
\bottomrule
\end{tabular}
\end{table}

After quality filtering and mapping to our clinical workflow taxonomy (screening-stage BI-RADS triage and diagnosis-stage lesion characterization), the BUS subset of BreastStage contains 10{,}405 unique image files and 190{,}730 instruction-following pairs (Table~\ref{tab:source_summary}), with expert-verified BI-RADS scores, lesion bounding boxes, and chain-of-thought rationales.

\paragraph{Mammography.}
Digital mammography is sourced from a subset of \textbf{MammoVQA}~\cite{mammovqa2025}, which originally unifies 15 public mammogram datasets. We adopt the 11 datasets that are licensed for derivative works and have stage-aware annotations compatible with our taxonomy (BMCD, CBIS-DDSM, CDD-CESM, CSAW-M, DMID, EMBED, INbreast, KAU-BCMD, MIAS, RSNA, VinDr-Mammo); we then re-organise their labels around the screening / diagnosis / treatment continuum and add new task templates on top of MammoVQA. Table~\ref{tab:mammo_internal} lists every adopted dataset together with its sample count and the task templates it supports. The union of these 11 sources covers the 9 clinically validated MammoVQA task templates—View (CC/MLO), Laterality, BI-RADS, Pathology, Masking potential, Background tissue, Subtlety, Density, and Abnormality detection (calcification, architectural distortion, mass, asymmetry, etc.).

\begin{table}[H]
\centering
\caption{Internal mammography source datasets aggregated by MammoVQA~\cite{mammovqa2025} and adopted as the BreastStage mammography source. ``Type'' follows the MammoVQA labelling: \textit{breast} = per-image labels, \textit{finding} = per-bounding-box labels, \textit{exam} = per-examination labels.}
\label{tab:mammo_internal}
\small
\begin{tabular}{l l r p{6cm}}
\toprule
\textbf{Dataset} & \textbf{Type} & \textbf{Size} & \textbf{Supported Tasks} \\
\midrule
BMCD~\cite{loizidou2022bmcd}            & breast  & 400    & Density, BI-RADS, Laterality \\
\multirow{2}{*}{CBIS-DDSM~\cite{lee2017cbisddsm}} & breast  & 3{,}103  & Density, BI-RADS, Laterality, View \\
                                                  & finding & 6{,}464  & Abnormality, Pathology, Subtlety \\
CDD-CESM~\cite{khaled2022cddcesm}        & breast  & 1{,}003  & Density, BI-RADS, Laterality, Pathology \\
\multirow{2}{*}{DMID~\cite{oza2024dmid}}  & breast  & 510    & Abnormality, Background tissue, Laterality, Pathology, View \\
                                          & finding & 868    & Abnormality, Pathology \\
INbreast~\cite{moreira2012inbreast}       & breast  & 410    & Density, Abnormality, BI-RADS, Laterality, View \\
\multirow{2}{*}{MIAS~\cite{suckling1994mias}} & breast  & 322    & Abnormality, Background tissue, Pathology \\
                                              & finding & 234    & Abnormality, Pathology \\
CSAW-M~\cite{sorkhei2021csawm}            & breast  & 10{,}020 & Laterality, Masking potential \\
KAU-BCMD~\cite{alsolami2021kaubcmd}       & breast  & 2{,}370  & BI-RADS \\
\multirow{2}{*}{VinDr-Mammo~\cite{nguyen2023vindrmammo}} & breast  & 20{,}000 & Density, Abnormality, BI-RADS, Laterality, View \\
                                                         & finding & 4{,}505  & Abnormality \\
RSNA~\cite{rsna2023}                      & breast  & 54{,}705 & Density, BI-RADS, Laterality, View \\
EMBED~\cite{jeong2023embed}               & exam    & 72{,}518 & Density (exam), BI-RADS (exam) \\
\midrule
\textbf{Internal total}                   & ---     & \textbf{177{,}432} & --- \\
\bottomrule
\end{tabular}
\end{table}

For BreastStage we use these 11 datasets with patient-level stratification preserved across train and test splits. After mapping the original classification labels to our screening / diagnosis / treatment stage taxonomy and synthesising stage-specific QA pairs, the mammography subset comprises 592{,}470 unique 2D images and 453{,}956 instruction-following pairs.

\paragraph{MRI.}
Multiparametric MRI sequences (T1-weighted, T2-weighted, DWI, DCE-MRI / T1dyn) are acquired from two collaborating clinical institutions under IRB approval. Board-certified breast radiologists at the contributing institutions provide expert annotations for tumor segmentation, molecular subtype prediction, and neoadjuvant chemotherapy response assessment. The MRI subset comprises 36{,}124 unique multiparametric sequence volumes (T1, T1dyn, T2W, DWI, ADC) across screening, diagnosis, and treatment-stage tasks, yielding 926{,}521 instruction-following pairs.

\paragraph{Whole Slide Imaging (WSI).}
Histopathology WSIs are sourced from three public repositories—\textbf{BCNB}, \textbf{TCGA-BRCA}, and \textbf{TCGA-HISTAI}—covering invasive ductal carcinoma, invasive lobular carcinoma, ductal carcinoma in situ, and benign / atypical lesions across histological grades I--III. Slides are scanned at 20$\times$ or 40$\times$ magnification with native resolutions ranging from 80{,}000$\times$80{,}000 to 120{,}000$\times$120{,}000 pixels. The WSI subset contains 2{,}510 slides with pathologist-verified molecular subtype labels (ER/PR/HER2 status, Ki-67 index, Nottingham grade), yielding 30{,}073 instruction-following pairs (captioning, closed- and open-ended VQA, plus prognosis / surgical-planning / systemic-therapy treatment-stage tasks).

\paragraph{Aggregate Statistics.}
Table~\ref{tab:source_summary} summarizes the per-modality data sources adopted by BreastStage. These modality-level counts correspond to the curated image pools before conversion into instruction-following examples. The final instruction corpus expands these image pools into multiple task-specific QA pairs, captions, grounding targets, and report-generation instances according to the stage-aware taxonomy described below.

\begin{table}[H]
\centering
\caption{Summary of BreastStage data sources by modality. ``Unique 2D / 3D images'' counts the distinct image (2D radiology, mammography, histopathology slide) and 3D-volume (CT, MRI multiparametric sequence) files used by the model, after filtering each modality to its retained sub-datasets. Each unique image typically backs many QA pairs.}
\label{tab:source_summary}
\small
\resizebox{\textwidth}{!}{%
\begin{tabular}{l p{4.6cm} p{3.4cm} r r}
\toprule
\textbf{Modality} & \textbf{Primary Source} & \textbf{Selection Criterion} & \textbf{Unique 2D / 3D Images} & \textbf{Instruction Pairs} \\
\midrule
CT      & CT-RATE~\cite{ctrate2024} & Female-breast volumes, post-QC & 20{,}546 volumes & 254{,}041 \\
BUS     & BUS-CoT~\cite{buscot2025} & All 99 histopathology cats & 10{,}405 images & 190{,}730 \\
Mammo   & MammoVQA-derived 11 sets~\cite{mammovqa2025}: BMCD, CBIS-DDSM, CDD-CESM, CSAW-M, DMID, EMBED, INbreast, KAU-BCMD, MIAS, RSNA, VinDr & Patient-level stratified, stage-specific re-organisation with new tasks added on top of MammoVQA & 592{,}470 images & 454{,}590 \\
MRI     & In-house multi-institution multiparametric MRI & Multiparametric studies with complete sequences & 36{,}124 volumes & 926{,}634 \\
WSI     & BCNB / TCGA-BRCA / TCGA-HISTAI & Subtype-labeled, 20--40$\times$ slides & 2{,}510 slides & 30{,}073 \\
\midrule
\textbf{Total} & \textbf{17 sub-datasets} & --- & \textbf{662{,}055 images} & \textbf{1{,}856{,}068} \\
\bottomrule
\end{tabular}%
}
\end{table}

\subsection{Instruction Data Generation Pipeline}\label{supp:instruction_pipeline}

\begin{figure}[H]
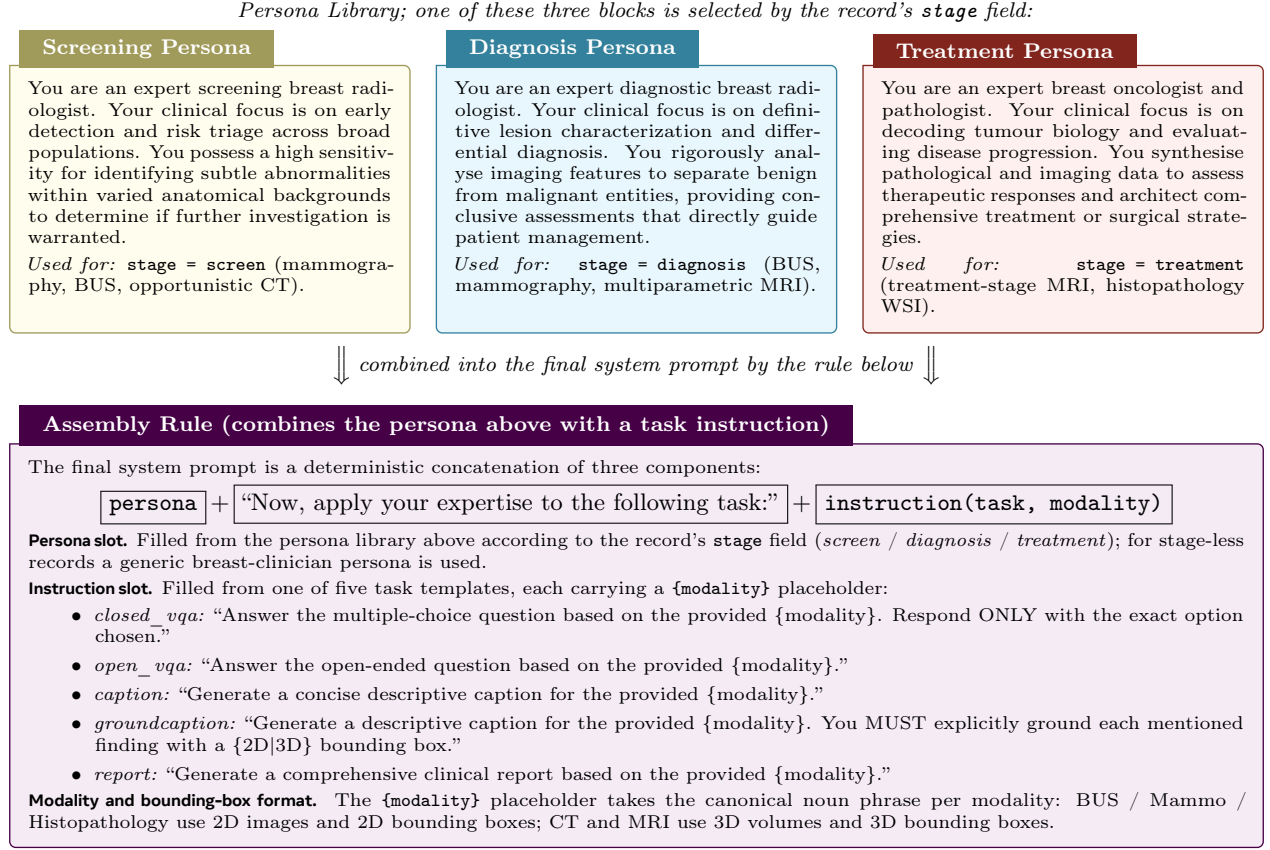

\centering
\textit{\footnotesize Persona Library; one of these three blocks is selected by the record's \texttt{stage} field:}\\[2pt]
\begin{minipage}[t]{0.32\linewidth}
\begin{personabox}[Screening Persona]{yellow}
You are an expert screening breast radiologist. Your clinical focus is on early detection and risk triage across broad populations. You possess a high sensitivity for identifying subtle abnormalities within varied anatomical backgrounds to determine if further investigation is warranted.

\textit{Used for:} \texttt{stage = screen} (mammography, BUS, opportunistic CT).
\end{personabox}
\end{minipage}\hfill
\begin{minipage}[t]{0.32\linewidth}
\begin{personabox}[Diagnosis Persona]{cyan}
You are an expert diagnostic breast radiologist. Your clinical focus is on definitive lesion characterization and differential diagnosis. You rigorously analyse imaging features to separate benign from malignant entities, providing conclusive assessments that directly guide patient management.

\textit{Used for:} \texttt{stage = diagnosis} (BUS, mammography, multiparametric MRI).
\end{personabox}
\end{minipage}\hfill
\begin{minipage}[t]{0.32\linewidth}
\begin{personabox}[Treatment Persona]{red}
You are an expert breast oncologist and pathologist. Your clinical focus is on decoding tumour biology and evaluating disease progression. You synthesise pathological and imaging data to assess therapeutic responses and architect comprehensive treatment or surgical strategies.

\textit{Used for:} \texttt{stage = treatment} (treatment-stage MRI, histopathology WSI).
\end{personabox}
\end{minipage}

\vspace{2pt}
\centerline{\footnotesize $\Big\Downarrow$ \emph{combined into the final system prompt by the rule below} $\Big\Downarrow$}
\vspace{2pt}

\begin{promptbox}[Assembly Rule (combines the persona above with a task instruction)]{violet}
The final system prompt is a deterministic concatenation of three components:
\begin{center}\small
\fbox{\texttt{persona}}\,$+$\,\fbox{``Now, apply your expertise to the following task:''}\,$+$\,\fbox{\texttt{instruction(task, modality)}}
\end{center}

\textbf{Persona slot.} Filled from the persona library above according to the record's \texttt{stage} field (\textit{screen} / \textit{diagnosis} / \textit{treatment}); for stage-less records a generic breast-clinician persona is used.

\textbf{Instruction slot.} Filled from one of five task templates, each carrying a \texttt{\{modality\}} placeholder:
\begin{itemize}
\setlength{\itemsep}{1pt}
\item \textit{closed\_vqa:} ``Answer the multiple-choice question based on the provided \{modality\}. Respond ONLY with the exact option chosen.''
\item \textit{open\_vqa:} ``Answer the open-ended question based on the provided \{modality\}.''
\item \textit{caption:} ``Generate a concise descriptive caption for the provided \{modality\}.''
\item \textit{groundcaption:} ``Generate a descriptive caption for the provided \{modality\}. You MUST explicitly ground each mentioned finding with a \{2D$|$3D\} bounding box.''
\item \textit{report:} ``Generate a comprehensive clinical report based on the provided \{modality\}.''
\end{itemize}

\textbf{Modality and bounding-box format.} The \texttt{\{modality\}} placeholder takes the canonical noun phrase per modality: BUS / Mammo / Histopathology use 2D images and 2D bounding boxes; CT and MRI use 3D volumes and 3D bounding boxes.
\end{promptbox}
\caption{Stage-aware system-prompt library used by BreastGPT. The three personas (top) cover the screening / diagnosis / treatment workflow; the assembly rule (bottom) deterministically composes a persona, a transition phrase, and a modality-specific task instruction at training and inference time. The same persona is reused regardless of modality, so a single model can adopt stage-appropriate reasoning style without task-specific heads.}
\label{fig:role_prompts}
\end{figure}

This section expands the four-stage curation pipeline introduced in Section~2 of the main paper (\textit{Workflow Data Generation} $\rightarrow$ \textit{Stage Instruction Data Construction} $\rightarrow$ \textit{Textual Description Generation} $\rightarrow$ \textit{Data Splitting and Verification}, illustrated in Figure~\ref{fig:dataset pipeline}). The four stages are powered by a Data Orchestration Engine that uses two LLMs in complementary roles: \textbf{Qwen2.5-VL-72B}~\cite{bai2025qwen25vl} is used wherever a decision requires looking at the image (modality-specific quality control, visual-attribute extraction), and \textbf{Qwen3-Max} is used for every text-only transformation (Chinese-to-English report parsing, structured-record-to-open-ended-VQA rewriting, ground-caption synthesis, report generation). Both LLMs are queried with strict JSON output schemas so downstream stages consume structured outputs without bespoke parsing. Per-modality QC prompts and the Chinese-to-English parsers appear in Figure~\ref{fig:prompt_templates}, and the stage-aware persona library together with the system-prompt assembly rule appear in Figure~\ref{fig:role_prompts}. The remainder of this section walks through the four stages in one-to-one correspondence with the main-paper subsections.

\paragraph{Workflow Data Generation.}
The cohort consists of public BUS, mammography, histopathology, and CT data plus an in-house multi-institution MRI dataset acquired from two collaborating hospitals under IRB approval; identifying information is removed at the source institutions before any data leaves the hospitals (\cref{supp:ethics_curation}). On top of the raw studies, this stage produces three artefacts that the rest of the pipeline relies on: per-image quality scores, lesion-level bounding-box / mask annotations, and stage-aware task labels.

\textit{Modality-specific quality control.} The visual specialist agents from Section~2 are implemented as Qwen2.5-VL-72B prompted with modality-specific quality-control rules (\cref{fig:prompt_templates}). The BUS selector rejects convex / sector-probe images outside the breast-ultrasound depth range; the mammography selector strictly rejects chest X-rays and applies PGMI positioning checks plus Eklund implant-displaced views; the breast MRI selector requires bilateral-coil framing or sagittal / axial single-breast views and verifies fat-saturation; the CT selector requires breast-inclusive FOV. Histopathology WSIs are sourced from a curated public collection where modality and magnification are already standardised, so they bypass this LLM-driven QC step. The selector emits a JSON \texttt{\{validity, reason\}}; samples flagged as low-quality are not silently dropped but routed to a \emph{Low-Image Check} step in which a breast specialist reviews the image and the LLM's stated reason, and either confirms the rejection or restores the sample.

\textit{Bounding-box and mask generation.} The four imaging modalities (BUS, mammography, CT, MRI) carry lesion-level spatial annotations, with the source differing by modality. BUS uses the expert hand-drawn lesion masks released with the BUS-CoT dataset~\cite{buscot2025}. Mammography (EMBED subset) ships with official lesion bounding boxes; we re-format the released \texttt{[ymin, xmin, ymax, xmax]} tuples into the \texttt{[xmin, ymin, xmax, ymax]} convention used throughout BreastStage, and no mask step is performed on this modality. CT lesion masks come from DRT-M3D~\cite{zhou2026dualres}, which also emits a per-volume cancer-risk score that we retain in the record metadata for the screening-stage CT tasks. MRI lesion masks are drawn manually on T1 and T1dyn sequences by 10 board-certified breast specialists at the contributing hospitals. For the mask-based modalities (BUS, CT, MRI), bounding boxes are derived by connected-component analysis: components with fewer than 50 voxels are discarded as noise, and the tight axis-aligned envelope of every retained component is emitted as a 4-tuple \texttt{[xmin, ymin, xmax, ymax]} for BUS or a 6-tuple \texttt{[xmin, ymin, zmin, xmax, ymax, zmax]} for the 3D volumes (CT, MRI). Histopathology WSIs do not carry pixel-level masks or bounding boxes in BreastStage; they are routed only to the closed/open VQA and report-generation tasks, not to the ground-caption task.

\textit{Expert-driven sub-task labelling.} After QC and mask generation, the per-dataset clinical tables and reports are reviewed by breast specialists who tag each record with one of \textbf{136 sub-task labels} reflecting BI-RADS items, mass / calcification descriptors, biomarker fields, and treatment-stage decisions. Each label is a \texttt{task template}: a high-level sub-task such as ``mass margin description'' or ``HER2 status'', typically spanning several leaf fields of the structured report defined in the next stage. Per-modality counts after this stage are reported in Supplementary Table~\ref{tab:source_summary}, and the resulting 136-task taxonomy in Table~\ref{tab:task_taxonomy_detailed}.

\paragraph{Stage Instruction Data Construction.}
This stage converts each curated study into a schema-bounded structured record, so that no free-text generation step downstream is allowed to introduce clinical content beyond what this record already encodes. The flow follows the four steps introduced in the main paper. (i) \textbf{Report translation.} Chinese radiology and pathology reports are translated to English by Qwen3-Max acting as a radiologist agent, using the bilingual prompts in Figure~\ref{fig:prompt_templates}. (ii) \textbf{Structured report extraction.} Every translated report is then parsed by the radiologist agent into an expert-designed BI-RADS-aligned structured report whose schema is written per modality by breast specialists following clinical reporting guidelines. For example, the MRI structured report (Figure~\ref{fig:mri_template}) covers per-side breast-level findings (FGT, BPE, post-surgical changes, lymph nodes, BI-RADS, management) plus per-lesion descriptors for masses, non-mass enhancement, and non-enhancing lesions. (iii) \textbf{Question bank derivation.} For each \texttt{task template} from Stage 1, breast specialists then write a closed-ended question that draws its answer from one or more leaf fields of the structured report, with options taken directly from the schema's enum values; this constrains BreastStage to questions that the structured record already answers. The resulting tuples carry the form \texttt{<Task, Question, options, Answer, Images>}. (iv) \textbf{Task--Stage mapping.} Specialists finally map each \texttt{task template} to one of the three clinical stages (\textit{screening} / \textit{diagnosis} / \textit{treatment}), which is also the key used for task-key balanced sampling on BreastStage-Bench; the per-stage record counts of the final task taxonomy are reported in Table~\ref{tab:task_taxonomy_detailed}.

From the structured records, BreastStage derives two parallel VQA partitions, both grounded in the same schema-bounded source. The \emph{closed-ended} partition packs the schema's enum values as an option list into the question text, so each tuple takes the form \texttt{<Task, Question+options, Answer, Images>} with the answer being one of the options. The \emph{open-ended} partition takes the same structured tuple, drops the option list, and asks Qwen3-Max to rewrite the categorical answer into a fluent clinical description, yielding \texttt{<Task, Question, Answer, Images>} where the answer is free-form text but the underlying clinical fact is unchanged; the rewriting prompt is reproduced in \cref{fig:struct_to_open_prompt}. Because both partitions are pinned to the same structured record, the LLM cannot fabricate clinical content beyond the schema, and the answer cannot leak through the question text into a text-only shortcut --- a property reflected in the $k{=}1$ collapse of BreastGPT's closed-ended accuracy in \cref{fig:token_budget}a, where the model degrades sharply once the visual evidence is removed.

\begin{figure}[H]
\centering
\begin{promptbox}[Open-VQA Generator]{teal}
You are a medical imaging data assistant specialising in breast \texttt{\{modality\}}. Your task is to rewrite a structured attribute into a natural-language Question--Answer pair suitable for a Breast \texttt{\{modality\}} VQA dataset.

\textbf{Rules.}
\begin{enumerate}
\setlength{\itemsep}{1pt}
\item The \texttt{Question} \emph{must} be open-ended and descriptive: do \emph{not} ask yes/no questions, do \emph{not} include options or binary choices.
\item The \texttt{Answer} \emph{must} be a complete sentence that is strictly based on the provided \texttt{Answer} value, and does \emph{not} introduce any new medical interpretation, diagnosis, or inference.
\item Do \emph{not} introduce information not explicitly contained in the input.
\item Do \emph{not} repeat or list the provided options in the output.
\item Use varied sentence structures across different samples.
\item Use precise anatomical and imaging terminology appropriate for breast \texttt{\{modality\}}.
\item Output \emph{only} valid JSON in the specified format. No extra text.
\end{enumerate}

\textbf{Input (JSON):} \texttt{\{"Question": "...", "options": ["...","...","..."], "Answer": "..."\}}.

\textbf{Output (JSON):} \texttt{\{"Question": "...", "Answer": "..."\}}.
\end{promptbox}

\vspace{4pt}
\begin{promptbox}[Grounded Caption Generator]{orange}
You are an \textbf{expert Radiologist Assistant} producing a \textbf{Grounded Caption} for breast \texttt{\{modality\}}. Strictly separate \emph{visual attributes} (shape, margin, echo / signal / density, posterior or kinetic features) from \emph{clinical context} (BI-RADS, risk, diagnosis, management).

\textbf{Step 1 -- Reference phrase.} For each finding, write a noun phrase that names \emph{only what is visible inside the bounding box}: include side, view (when applicable), abnormality type, and the relevant visual descriptors. Exclude BI-RADS, risk, diagnosis, and management.

\textbf{Step 2 -- Caption skeleton.} Compose a professional radiology sentence with placeholder tokens \texttt{<ref-object><bbox>} immediately after each lesion phrase; inject the excluded clinical context (BI-RADS, risk, recommendation) \emph{around} the placeholders, never inside the reference phrase.

\textbf{Step 3 -- Strict alignment.} Emit \texttt{ref[]}, \texttt{bbox[]} as flat lists in caption order; \texttt{ref[i]} corresponds to \texttt{bbox[i]} and to the i-th \texttt{<ref-object><bbox>} token. Boxes are 2D \texttt{[xmin,ymin,xmax,ymax]} for BUS / mammography / histopathology and 3D \texttt{[xmin,ymin,zmin,xmax,ymax,zmax]} for CT / MRI.

\textit{Output JSON:} \texttt{\{caption, objects:\{ref, bbox\}\}}.
\end{promptbox}

\vspace{4pt}
\begin{promptbox}[Report Generator]{purple}
You are a \textbf{board-certified breast radiologist}. Generate a formal breast \texttt{\{modality\}} report from the structured \texttt{patient\_data}, compliant with the \textbf{ACR BI-RADS Lexicon} for that modality and reflecting real-world dictation style.

\textbf{Hard safety rules.} (1) No fabrication of laterality, lesion size, clock-face position, or distance to nipple unless explicitly present in \texttt{patient\_data}; if missing, refer generically (``the lesion''). (2) No histological / molecular confirmation: replace ``X is invasive carcinoma'' with probabilistic phrasing (``findings are suspicious for X''). (3) Bounding boxes are internal; never mention coordinates, ``bbox'', or any data-structure terminology in the prose.

\textbf{Sections (fixed order).} \emph{Findings}: one passive-voice paragraph per lesion, ordered by descending BI-RADS, integrating only descriptors present in \texttt{patient\_data}. \emph{Impression}: probabilistic synthesis. \emph{Final Assessment}: the highest BI-RADS category among findings. \emph{Management}: ACR-aligned recommendation (Cat 2 -- routine; Cat 3 -- short-interval follow-up; Cat 4--5 -- biopsy / tissue diagnosis). Surgical / systemic therapy decisions are out of scope.

\textit{Output:} four-section markdown with the section names above.
\end{promptbox}
\caption{Generator prompts that produce the BreastStage instruction-following partitions from each structured record: the open-ended VQA generator (top), the grounded caption generator (middle), and the report generator (bottom). All three are instantiated per modality with \texttt{\{modality\}} $\in$ \{BUS, mammography, CT, MRI, histopathology\}.}
\label{fig:struct_to_open_prompt}
\end{figure}

\begin{figure}[H]
\centering
\begin{promptbox}[MRI Structured Report Template]{violet}
\begin{lstlisting}[language=jsonsch,basicstyle=\fontsize{7}{7}\ttfamily,aboveskip=0pt,belowskip=0pt]
{
  "breast_level": {
    "FGT": {"left": "C", "right": "C"},                    // "A","B","C","D"
    "BPE": {"left": "Moderate", "right": "Mild"},          // "Minimal","Mild","Moderate","Marked"
    "breast_symmetry": true,
    "post_surgical_changes": {
      "left":  {"scar_tissue": false, "lumpectomy_changes": false, "mastectomy_changes": false},
      "right": {"scar_tissue": false, "lumpectomy_changes": false, "mastectomy_changes": false}
    },
    "associated_features": {
      "left": {
        "nipple_retraction": null, "nipple_invasion": null,
        "skin_retraction": null, "skin_thickening": null,
        "skin_invasion": {"direct_invasion": null, "inflammatory_cancer": null},
        "axillary_adenopathy": null, "pectoralis_muscle_invasion": null,
        "chest_wall_invasion": null, "architectural_distortion": null,
        "ductal_dilation": null                            // "mild","marked",null
      },
      "right": { ... same schema as "left" ... }
    },
    "lymph_nodes": {
      "axillary_left":  null, "axillary_right":  null,     // "normal","abnormal"
      "internal_mammary_left": null, "internal_mammary_right": null
    },
    "BI-RADS":   {"left": null, "right": null},            // "0","1","2","3","4A","4B","4C","5","6"
    "management":{"follow_up_recommended": null, "biopsy_recommended": null}
  },
  "mass": [
    {
      "id": 1, "side": "left",
      "location": {"quadrant": "UOQ",                      // "UIQ","UOQ","LIQ","LOQ","central",null
                   "depth":    "middle"},                  // "anterior","middle","posterior",null
      "size_mm":  {"long": 12, "short": 9, "depth": 8},
      "signal_characteristics": {
        "T1W": "hyperintense", "T2W": "hyperintense",      // "hyperintense","isointense","hypointense",null
        "DWI_restriction": "present",                      // "present","absent",null
        "ADC": "hyperintense",
        "kinetics": {"initial": "Fast",                    // "Slow","Medium","Fast",null
                     "delayed": "Wash-out"}                // "Persistent","Plateau","Wash-out",null
      },
      "morphology": {"shape": "oval",                      // "round","oval","lobular","irregular",null
                     "margin": "circumscribed"},           // "circumscribed","irregular","spiculated",null
      "internal_enhancement": {
        "pattern": "heterogeneous",                        // "homogeneous","heterogeneous","rim_enhancement","dark_internal_septations",null
        "associated_vascularity": {"adjacent_vessel_sign": null,
                                   "increased_peritumoral_vascularity": null}
      }
    }
  ],
  "non_mass_enhancement": [
    {
      "lesion_id": 1, "side": "left",
      "location": {"quadrant": null, "depth": null},
      "size_mm":  {"long": null, "short": null, "depth": null},
      "distribution": "Segmental",                         // "Focal","Linear","Regional","Segmental","Multiple regions","Diffuse",null
      "internal_enhancement_pattern": "Clustered_ring",    // "Homogeneous","Heterogeneous","Clustered","Clustered_ring",null
      "signal_characteristics": { ... same as "mass" ... }
    }
  ],
  "non_enhancing_lesion": [
    {
      "lesion_id": 1, "side": "left",
      "location": {"quadrant": null, "depth": null},
      "size_mm":  {"long": null, "short": null, "depth": null},
      "signal_characteristics": { ... same as "mass" ... },
      "lesion_type_detail": "ductal_precontrast_high_signal_on_T1W"
                            // "cyst","non_enhancing_mass","architectural_distortion",null
    }
  ]
}
\end{lstlisting}
\end{promptbox}
\caption{Expert-designed BI-RADS-aligned structured template used for the MRI cohort. Every leaf field is either a numeric measurement, a Boolean flag, or one of a fixed enum list (shown as a comment). Reports are first translated to English by Qwen3-Max and then populated into this template under its enum and type constraints; specialists then derive a question per \texttt{task template} whose answer is drawn from one or more leaf fields. Schema-conformant population guarantees that the resulting QA pairs cannot introduce facts that the source report does not already record.}
\label{fig:mri_template}
\end{figure}

\paragraph{Textual Description Generation.}
The same structured records that feed VQA construction are also reused to synthesise the two narrative task formats, ground caption and report generation, with the modality-parameterised prompts in Figure~\ref{fig:struct_to_open_prompt}. Both formats remain pinned to the structured record, so Qwen3-Max is responsible only for the surface form and never for the underlying clinical fact.

\textit{Ground caption.} Ground captions are produced only for the four imaging modalities that carry bounding boxes (BUS, mammography, CT, MRI); histopathology, which has no boxes, is excluded. Each caption is generated by binding every visible lesion to its box and weaving the resulting visually grounded phrases into a clinical narrative; the surrounding context drawn from the structured record (BI-RADS, risk level, recommendation) is allowed to appear around the boxes but never inside the lesion phrase, so a reader can recover, from the caption alone, which descriptors apply to which spatial region. The modality-specific vocabularies differ in the obvious places (posterior acoustic features for BUS, ACR density for mammography, FGT, BPE and dynamic kinetic curves for MRI), but the visual-vs-context separation is identical across modalities.

\textit{Report generation.} The report generator targets all five modalities, with one prompt variant per modality so that the output respects the relevant ACR BI-RADS lexicon. Every report is forced into the same four sections (Findings, Impression, Final Assessment, Management) so that downstream evaluation can score them section by section. Three constraints prevent the failure modes typical of pure-LLM report generators: spatial details that are not in the structured input (laterality, lesion size, clock-face position, distance to nipple) cannot appear, source pathology labels must remain probabilistic in the prose (``findings are suspicious for invasive carcinoma'' rather than ``this is invasive carcinoma''), and any reference to internal field names or coordinates is forbidden so the report reads as clinical dictation. Drafts then flow into the specialist-audit loop in Stage 4 below, where modality-specific prompts are revised and failing records regenerated whenever a per-task flag rate exceeds threshold.

\paragraph{Data Splitting and Verification.}
Records are split between training and BreastStage-Bench at patient-id granularity; because the source datasets cover disjoint patient populations, no cross-dataset patient deduplication is required. The split itself uses a single global stratified sampler over the composite (modality, task type, pathology label) key, so the resulting train and test partitions match on the underlying task and class distributions. BreastStage-Bench equals the test partition, drawn with task-key balanced sampling so that every task template in Table~\ref{tab:task_taxonomy_detailed} is represented. Verification has two layers. Automated heuristics first remove records with malformed bounding-box coordinates, invalid option labels, instruction--answer conflicts, hallucinated clinical terms, and near-duplicate QA pairs (MinHash $> 0.85$). Three breast specialists then conduct an independent task-level audit on Bench: each specialist reviews 5 random samples per task and flags clinically inconsistent or stage-mismatched cases. Any task whose flag rate is non-negligible across specialists triggers re-running the relevant LLM stage with the failing prompts revised.

\begin{figure}[H]
\centering
\begin{minipage}[t]{0.49\linewidth}
\begin{promptboxA}[BUS Quality Selector]{yellow}
You are a \textbf{Breast Ultrasound Quality Specialist}. Your goal is to validate if the image is likely a \textbf{Breast/Axilla Ultrasound}. Evaluate the following:

\textbf{1. Modality \& Probe Geometry (BREAST FOCUS):}
\begin{itemize}\setlength\itemsep{0pt}\setlength\topsep{0pt}
\item \textit{Linear Probe (Rectangular Field):} high probability of valid Breast US.
\item \textit{Convex/Sector Probe (Fan-shaped Field):} \textbf{WARNING} -- typical for Abdomen/Liver. Mark Valid only if the text explicitly confirms a deep breast mass.
\item \textit{Modality:} confirm Ultrasound (speckle texture).
\end{itemize}

\textbf{2. Image Quality \& Settings:}
\begin{itemize}\setlength\itemsep{0pt}\setlength\topsep{0pt}
\item \textit{Depth:} $<$ 4--6\,cm; reject if deep abdominal structures (liver / kidney) are visible.
\item \textit{Focus zone:} aligned with the mammary layer.
\item \textit{Gain:} fat / glandular interface distinguishable.
\end{itemize}

\textbf{3. Text--Image Alignment:} flag if the text mentions ``Liver'', ``Gallbladder'', or ``Thyroid''.

\textit{Output JSON:} \{validity: Valid/Invalid, reason\}.
\end{promptboxA}
\end{minipage}\hfill
\begin{minipage}[t]{0.49\linewidth}
\begin{promptboxA}[Mammography Quality Selector]{cyan}
You are a \textbf{Mammography Quality Control Expert}. Validate if the image is a diagnostic \textbf{Mammogram (MG)}. Evaluate the following:

\textbf{1. Modality Distinction (CRITICAL):}
\begin{itemize}\setlength\itemsep{0pt}\setlength\topsep{0pt}
\item Confirm Mammogram: soft-tissue breast against black air background.
\item \textbf{STRICTLY REJECT} Chest X-Ray (CXR): if you see ribs, lungs, clavicles, or the spine, mark Invalid.
\end{itemize}

\textbf{2. Breast-Specific Positioning (PGMI Criteria):}
\begin{itemize}\setlength\itemsep{0pt}\setlength\topsep{0pt}
\item \textit{Pectoralis muscle:} for MLO views, must extend to the nipple line.
\item \textit{Nipple profile:} in profile (not superimposed) unless retracted.
\item \textit{Skin folds:} reject severe folds that mimic abnormalities.
\item \textit{Implants:} identify ``Implant Displaced (Eklund)'' views vs.\ standard views.
\end{itemize}

\textbf{3. Exposure \& Contrast:} adipose (dark gray) vs.\ fibroglandular tissue (white) must be distinguishable; reject motion blur at calcifications.

\textbf{4. Text--Image Alignment:} text describes mammographic findings (asymmetry, distortion, calcification).

\textit{Output JSON:} \{validity, reason\}.
\end{promptboxA}
\end{minipage}

\vspace{4pt}
\begin{minipage}[t]{0.49\linewidth}
\begin{promptboxB}[Breast MRI Quality Selector]{red}
You are a \textbf{Breast MRI Physics Specialist}. Validate if the image is a diagnostic \textbf{Breast MRI} sequence. Evaluate the following:

\textbf{1. Coil \& Field of View (BREAST FOCUS):}
\begin{itemize}\setlength\itemsep{0pt}\setlength\topsep{0pt}
\item \textit{Breast coil geometry:} bilateral breasts hanging in the coil, or a dedicated sagittal/axial single-breast view.
\item \textit{Exclude:} Brain MRI, Spine MRI, or Abdominal MRI where the breast is incidental and compressed.
\item \textit{FOV:} must include the axillary tail region.
\end{itemize}

\textbf{2. Sequence \& Physics:}
\begin{itemize}\setlength\itemsep{0pt}\setlength\topsep{0pt}
\item \textit{Fat Saturation:} \textbf{CRITICAL} for breast -- fat must be suppressed (dark) except in T1 non-fat-sat anatomy scans.
\item \textit{Cardiac motion artefacts:} check the left-breast phase-encoding direction; severe pulsation rendering the medial breast unreadable $\rightarrow$ Invalid.
\item \textit{Silicone select:} for implant cases, confirm silicone-specific sequences are used.
\end{itemize}

\textbf{3. Text--Image Alignment:} text describes ``enhancement kinetics'', ``BPE'', or ``washout''; flag mismatch if it describes CT density.

\textit{Output JSON:} \{validity, reason\}.
\end{promptboxB}
\end{minipage}\hfill
\begin{minipage}[t]{0.49\linewidth}
\begin{promptboxB}[Side-Aware Report Splitter]{green!60!black}
You are an \textbf{expert breast radiologist}. Process Chinese radiology reports: (1) split findings by side, (2) translate everything into professional medical English, (3) determine benign / malignant from findings and BI-RADS.

\textbf{CRITICAL RULE -- NO CHINESE.} The final JSON output must not contain any Chinese characters; the thinking and the result are 100\% English.

\textbf{Side-assignment rules (per sentence):}
\begin{itemize}\setlength\itemsep{0pt}\setlength\topsep{0pt}
\item left-side keyword (\textit{zuo / zuoce / zuoru}) $\rightarrow$ LEFT.
\item right-side keyword (\textit{you / youce / youru}) $\rightarrow$ RIGHT.
\item bilateral / no side mentioned $\rightarrow$ include in BOTH, rewritten per side.
\item Sentences mentioning both sides (e.g.\ ``Left is smaller than right'') are split into two side-specific English statements.
\end{itemize}

\textbf{Translation standard:} use professional English medical terminology (``hyperplasia'', ``BI-RADS'', ``outer upper quadrant''); preserve every measurement, echo, and vascularity descriptor.

\textit{Output JSON:} \{left: \{finding, \dots\}, right: \{finding, \dots\}\}.
\end{promptboxB}
\end{minipage}

\vspace{4pt}
\begin{promptbox}[Bilingual Pathology Report Parser]{violet}
You are an \textbf{expert Pathologist and Data Structuring Specialist}. Extract information from raw Chinese breast pathology reports and convert it into a structured \textbf{English JSON} format.

\textbf{Input format.} \{\texttt{type}: ``Surgical pathology'' or ``Non-surgical pathology''; \texttt{pathology}: report text\}. Surgical pathology is text fields separated by spaces or colons. Non-surgical pathology is often a Python set string such as \texttt{"\{'text', 'text'\}"}.

\textbf{Symbol decoding.} \texttt{(-)} $\rightarrow$ Negative; \texttt{(+)} $\rightarrow$ Positive; \texttt{/} $\rightarrow$ Not Applicable or Not Found.

\textbf{Translation rules (strict adherence).} Histology terms map deterministically: \textit{jin run xing ai} $\rightarrow$ Invasive Carcinoma; \textit{fei te shu lei xing} (NST) $\rightarrow$ Invasive Carcinoma of No Special Type (NST); \textit{dao guan yuan wei ai} $\rightarrow$ Ductal Carcinoma In Situ (DCIS); \textit{xian wei xian liu} $\rightarrow$ Fibroadenoma; \textit{ying hua xing xian bing} $\rightarrow$ Sclerosing Adenosis; \textit{ru xian bing} $\rightarrow$ Mastopathy; \textit{suo xing xi bao} $\rightarrow$ Spindle Cell; \dots\ Biomarkers (ER, PR, HER2, Ki-67, Nottingham grade) and molecular subtype are normalized to the canonical Enum (Luminal A / B, HER2-enriched, TNBC).

\textit{Output JSON:} canonical pathology schema; missing fields left null.
\end{promptbox}
\caption{System prompts of the five BreastStage data-orchestration agents (modality-specific quality selectors plus the bilingual report splitter and pathology parser). Each agent runs Qwen3-Max with a strict JSON output schema, so its decision can be consumed by the next stage without manual parsing. Generator prompts that produce the actual instruction-following pairs are reproduced in Figure~\ref{fig:struct_to_open_prompt}.}
\label{fig:prompt_templates}
\end{figure}

\paragraph{Stage-aware system prompts.} BreastStage uses two complementary prompt families. The data-orchestration prompts in Figure~\ref{fig:prompt_templates} and Figure~\ref{fig:struct_to_open_prompt} drive the curation pipeline: the five orchestration agents (BUS / Mammography / Breast MRI Quality Selectors, Side-Aware Report Splitter, Bilingual Pathology Report Parser) and the three generators (Open-VQA, Grounded Caption, Report). At training and inference time, the resulting instruction-following pairs are then served with stage-specific system prompts that switch BreastGPT between three clinical roles: screening radiologist, diagnostic radiologist, and treatment-stage breast oncologist or pathologist; Figure~\ref{fig:role_prompts} reproduces the three personas and the deterministic template that combines a persona with a modality-aware task instruction.

\subsection{Task Taxonomy and Distribution}\label{supp:task_taxonomy}

\begin{figure}[H]
\centering
\includegraphics[width=\linewidth]{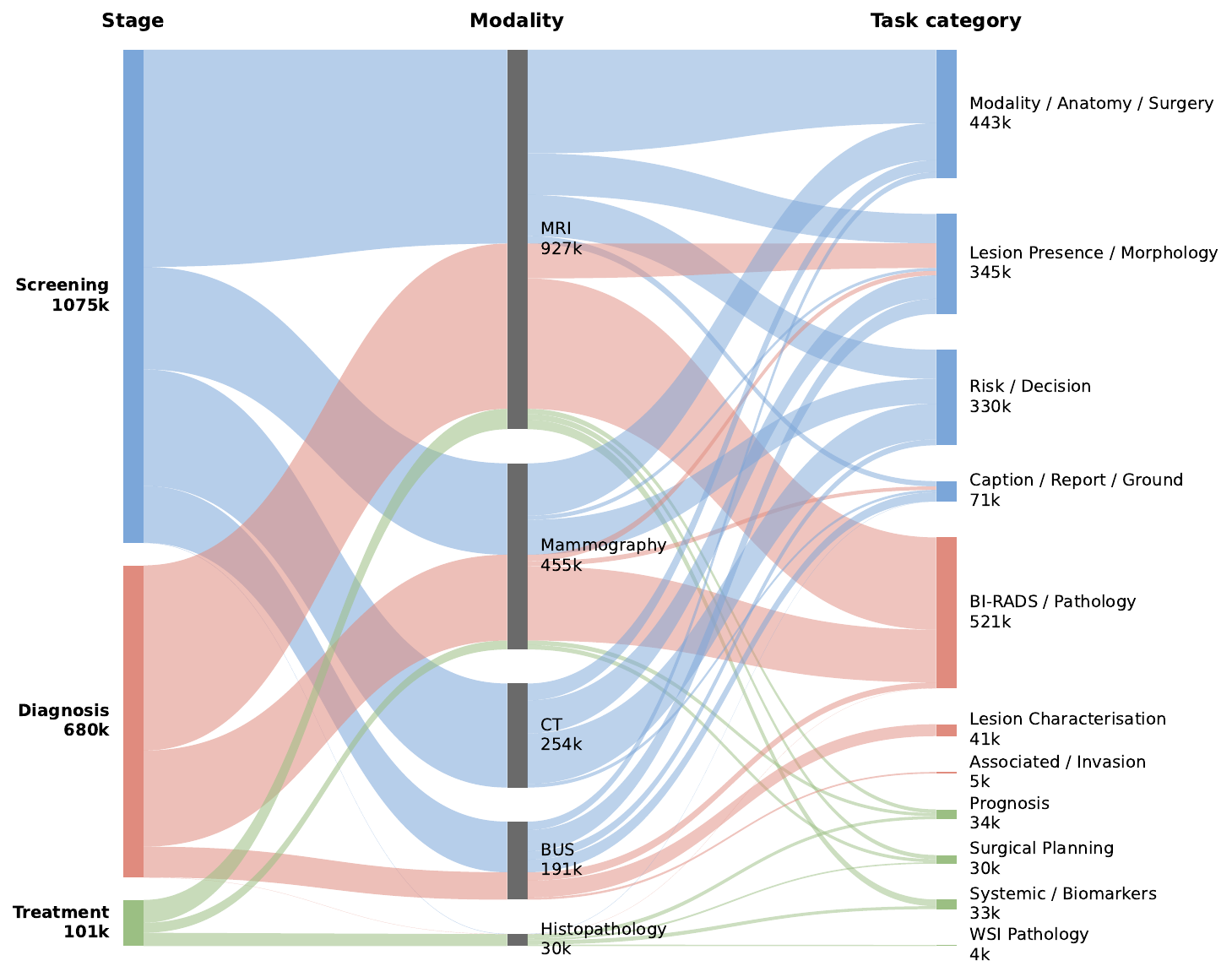}
\caption{Joint distribution of the 1.86M BreastStage instruction-following pairs across clinical stage (left), imaging modality (middle), and task category (right). Ribbon width is proportional to record count, and ribbons are coloured by clinical stage so each contributing stage remains identifiable on both sides of the modality node.}
\label{fig:stage_modality_task_sankey}
\end{figure}

We summarise the BreastStage instruction corpus along two complementary axes. Table~\ref{tab:task_taxonomy_detailed} groups the 136 task templates into a 12-category taxonomy under the three clinical stages, and Figure~\ref{fig:stage_modality_task_sankey} reports the joint distribution over (clinical stage, imaging modality, task category) that this taxonomy induces.

\begin{table}[H]
\centering
\caption{Twelve-category task taxonomy of BreastStage. Templates are grouped by clinical stage (Screening, Diagnosis, Treatment) and, within each stage, by the family of clinical concept being assessed. The \emph{Samples} column reports the number of instruction-following pairs whose majority clinical stage falls in that category.}
\label{tab:task_taxonomy_detailed}
\small
\resizebox{\textwidth}{!}{%
\begin{tabular}{p{2.2cm}|p{4.0cm}|p{6.5cm}|r}
\toprule
\textbf{Clinical Stage} & \textbf{Task Category} & \textbf{Representative Task Templates} & \textbf{Samples} \\
\midrule
\multirow{4}{2.2cm}{Screening\\(1{,}075{,}092)}
& Modality / Anatomy / Surgical History
  & Modality, View, Laterality, Orientation, ACR, FGT, BPE, Background tissue, breast\_symmetry, post\_surgical\_changes (mastectomy / lumpectomy / scar)
  & 442{,}860 \\
& Lesion Presence \& Morphology
  & abnormal\_presence, abnormality\_type, mass, non\_mass\_enhancement, non\_enhancing\_lesion, Shape, LesionEdge, LesionBoundary, EchoCharacteristics, LesionCalcificationFeatures, BloodFlow, lesion depth
  & 244{,}188 \\
& Risk Assessment \& Screening Decision
  & RiskLevel, ScreeningDecision, Density\_Risk, follow\_up\_recommended
  & 329{,}756 \\
& Captioning / Report / Grounding
  & ground caption (BUS, CT), report generation (BUS), Report (MRI)
  & 58{,}288 \\
\midrule
\multirow{4}{2.2cm}{Diagnosis\\(680{,}409)}
& BI-RADS \& Pathology Diagnosis
  & BI-RADS, birads, diagnosis, Pathology, main\_type
  & 520{,}969 \\
& Lesion Characterization
  & mass.morphology (margin, shape), mass.signal (T1W, T2W, ADC, DWI, kinetics), internal\_enhancement, calcificationType, boundaryDefinition, echoTexture, morphology, distribution
  & 141{,}761 \\
& Captioning / Report / Grounding (Mammo / MRI)
  & ground caption (Mammo), report generation (Mammography, MRI)
  & 12{,}227 \\
& Associated Findings \& Invasion
  & lymph\_nodes (axillary, internal\_mammary), skin\_thickening, nipple\_retraction, architectural\_distortion, ductal\_dilation, pectoralis / chest\_wall / nipple / skin invasion, vascularity, invasiveness
  & 5{,}452 \\
\midrule
\multirow{4}{2.2cm}{Treatment\\(100{,}567)}
& Prognosis \& Outcome Prediction
  & prognosis, Treatment.Prognosis, Treatment.treatment.prognosis
  & 33{,}743 \\
& Surgical Planning \& Urgency
  & surgical\_plan, SurgicalPlan, treatment.urgency
  & 29{,}797 \\
& Systemic Therapy \& Molecular Biomarkers
  & systemic\_therapy, biomarkers (ER, PR, HER2, Ki67, molecular\_subtype)
  & 33{,}200 \\
& WSI Pathology Subtyping \& Captioning
  & Tumor, ER, PR, HER2, HER2 Expression, Molecular subtype, Histological grading, slide caption / VQA
  & 3{,}827 \\
\bottomrule
\end{tabular}%
}
\end{table}

\section{BreastGPT Technical Details}\label{supp:breastgpt_details}

\subsection{Base Model Configuration}\label{supp:base_model}

BreastGPT is initialized from Qwen3-VL-8B-Instruct and inherits its instruction-following multimodal interface. We keep the language backbone and the standard radiology visual encoder largely aligned with the released model and introduce additional components only where the breast cancer workflow requires capabilities not naturally covered by the base model. The major architectural extension is the GigaPixel branch for WSI processing together with the coverage-maximizing token selector that maps heterogeneous image representations into a fixed downstream token budget. For CT, BUS, mammography, and MRI inputs, the standard branch is used; for WSI inputs, modality-aware routing activates the GigaPixel branch described below.

\subsection{GigaPixel Encoder Implementation}\label{supp:gigapixel}

The GigaPixel encoder is designed to preserve diagnostically relevant pathology patterns without forcing the LLM to attend over the full patch sequence. A single WSI can contain tens or hundreds of thousands of tiles, and many clinically important cues are sparse: mitotic figures, ductal structures, invasive fronts, necrosis, lymphocytic infiltration, and tumor-stroma interfaces may occupy only a small fraction of the slide. The WSI branch therefore separates representation learning into three steps: local patch encoding, long-range contextualization, and query-aware coverage selection.

\paragraph{Stage 1: Patch Extraction and Encoding.}
WSI tiles are extracted at 20$\times$ magnification with non-overlapping 512$\times$512 pixel patches. For a typical breast WSI this yields tens of thousands of patches; CONCH v1.5 patch features are pre-computed once per slide and stored, following the path convention \texttt{20x\_512px\_0px\_overlap/features\_conch\_v15/}. Each patch is encoded using the frozen CONCH v1.5 foundation model~\cite{anon2025visuallanguage}, producing 512-dimensional embeddings:
\[
\mathbf{v}_i = \text{CONCH}(\mathbf{x}_i), \quad i = 1, \ldots, N, \quad \mathbf{v}_i \in \mathbb{R}^{512}.
\]

\paragraph{Stage 2: LongNet Contextualization.}
The patch embeddings are processed by a 2-layer LongNet encoder with dilated attention (matching the configuration described in Section 3 of the main paper). The dilation rates follow an exponential schedule, allowing the model to capture both local cellular details and global tissue architecture. The contextualized embeddings are:
\[
\mathbf{h}_i = \text{LongNet}(\{\mathbf{v}_1, \ldots, \mathbf{v}_N\})_i.
\]
LongNet's exponentially growing dilation rates give the encoder long-range coverage without the $O(N^2)$ cost of full attention.

\subsection{Coverage-Maximizing Token Selection}
\label{subsec:mmtok}

The coverage-maximizing token selector is the mechanism that makes BreastGPT practical as a single model across ordinary radiology images and gigapixel pathology slides. Without compression, WSI inputs would exceed the LLM context budget by several orders of magnitude; with naive pooling or truncation, sparse but clinically important regions may be removed. We therefore use a coverage objective that keeps the selected tokens both \emph{clinically query-relevant} and \emph{globally representative} of the visual evidence.

Given a sequence of projected visual tokens and an associated clinical instruction, the goal of the selector is to identify a compact subset $S \subset \{1, \ldots, N\}$ with $|S| = k$ ($k=128$ by default) that jointly preserves relevance to the clinical query and fidelity to the overall visual distribution. For WSI inputs, the source sequence is the LongNet-contextualized patch representation $\{h_1, \ldots, h_N\}$; for standard radiological inputs, the source sequence is the ViT token sequence. In both cases, the selected tokens are passed to the LLM as the fixed-budget visual context. This gives BreastGPT a modality-invariant interface: each clinical image, regardless of its original resolution, is represented by the same number of visual tokens before language-model reasoning.

\paragraph{Similarity matrices.}
Let $\{v_1, \ldots, v_N\}$ denote the projected vision tokens after the LLM projection layer, aligned with the text token space, and let $\{v'_1, \ldots, v'_N\}$ denote the pre-projection tokens that retain modality-native similarity structure. Let $\{t_1, \ldots, t_m\}$ denote the text tokens encoding the clinical instruction. We define text--vision (T-V) and vision--vision (V-V) similarity matrices as
\begin{equation}
M^{tv}_{i,j} = t_i^\top v_j, \qquad
M^{vv}_{i,j} = {v'}_i^{\!\top} v'_j, \qquad
\|t_i\|_2 = \|v_j\|_2 = \|v'_j\|_2 = 1.
\end{equation}
The similarities are then softmax-calibrated with temperatures $\tau_t$ and $\tau_v$:
\begin{equation}
\widetilde{M}^{tv}_{i,j} =
\frac{\exp(M^{tv}_{i,j}/\tau_t)}
{\sum_{j'} \exp(M^{tv}_{i,j'}/\tau_t)}, \qquad
\widetilde{M}^{vv}_{i,j} =
\frac{\exp(M^{vv}_{i,j}/\tau_v)}
{\sum_{j'} \exp(M^{vv}_{i,j'}/\tau_v)}.
\end{equation}

\paragraph{Dual coverage objective.}
We select $S$ by maximizing the joint coverage score
\begin{equation}
f(S) =
\underbrace{\frac{1}{m}\sum_{i=1}^{m} \max_{s \in S} \widetilde{M}^{tv}_{i,s}}_{\text{T-V coverage}}
+ \alpha \cdot
\underbrace{\frac{1}{N}\sum_{i=1}^{N} \max_{s \in S} \widetilde{M}^{vv}_{i,s}}_{\text{V-V coverage}},
\end{equation}
where $\alpha$ balances query relevance against global visual representativeness. The T-V term rewards tokens that explain the clinical instruction, while the V-V term forces the selected subset to cover the full visual distribution so that diagnostically relevant but less immediately salient regions are not discarded.

\paragraph{Why both terms are needed.}
The two coverage terms play complementary roles. T-V coverage favors tokens aligned with the prompt, such as lesion regions for BI-RADS questions or tumor-cell regions for molecular subtype questions. However, relying on T-V coverage alone can over-concentrate the selected subset around a small number of highly salient regions. V-V coverage counteracts this collapse by selecting tokens that represent the broader slide or image distribution. This is especially important for pathology, where grade, subtype, and treatment-response cues can depend on tissue architecture, tumor-stroma interaction, and heterogeneity across multiple regions. Conversely, V-V coverage alone may preserve visually diverse background regions that are irrelevant to the current clinical task. The combined objective gives BreastGPT a principled way to balance prompt alignment and visual diversity.

\paragraph{Greedy optimization.}
Each component of $f$ is a facility-location function over the token set and is therefore monotone submodular; the non-negative linear combination remains monotone submodular. A standard greedy algorithm that iteratively adds the token with the largest marginal gain achieves the usual $(1-\nicefrac{1}{e})$ approximation to the NP-hard optimum in $O(kN)$ time. End-to-end measured selector latency at $k=128$ on a representative histopathology WSI is reported in Table~\ref{tab:inference_efficiency}.

\paragraph{Implementation details.}
The selector is applied after modality-specific visual encoding and before LLM injection. For standard radiology images, token selection operates on the ViT token sequence. For WSI inputs, token selection operates on LongNet-contextualized patch embeddings. The same value of $k$ is used for both branches, which ensures that radiology and pathology inputs have comparable downstream inference costs. During training, the selector is run on-the-fly so that the language model learns to reason over the same type of compact visual context used at inference time.


\subsection{Training Configuration}\label{supp:training_config}

BreastGPT is trained in two stages. In \textit{Stage 1}, we warm up the visual front end on the closed-VQA subset of BreastStage: the LLM backbone is frozen, while the ViT, the aligner (which includes the WSI projection and LongNet module), and the resolution gate are trainable; the CONCH v1.5 encoder is always frozen. This step aligns the visual representations, both the ViT branch for radiology and the CONCH+LongNet branch for WSIs, with the multimodal token space before any gradient flows into the language model. In \textit{Stage 2}, we unfreeze the LLM and continue end-to-end fine-tuning on all four task formats across every modality; the CONCH v1.5 encoder remains frozen. The coverage token selector is training-free and applied identically at training and inference. We use AdamW with a cosine learning rate schedule and DeepSpeed ZeRO-2, cap the per-image visual budget at 1024 tokens, cap CT and MRI volumes at $384\times 384\times 48$, and fix the selector budget at $k=128$ for both branches so that radiological and pathological inputs incur the same downstream inference cost. Table~\ref{tab:training_config} summarises the two-stage training configuration used for BreastGPT, taken directly from the launch scripts.

\begin{table}[H]
\centering
\caption{Training configuration for BreastGPT. Both stages use the \texttt{ms-swift} \texttt{sft} entry point with FlashAttention, gradient checkpointing, and DeepSpeed ZeRO-2.}
\label{tab:training_config}
\small
\resizebox{\linewidth}{!}{%
\begin{tabular}{p{3.1cm} p{4.9cm} p{4.9cm}}
\toprule
\textbf{Setting} & \textbf{Stage 1: Visual front-end warm-up} & \textbf{Stage 2: End-to-End SFT} \\
\midrule
Epochs & 1 & 2 \\
Trainable modules & ViT, aligner (incl.\ WSI projection / LongNet) & LLM, ViT, aligner \\
Frozen modules & LLM backbone, CONCH encoder & CONCH encoder \\
Datasets & Closed-VQA subsets (BUS, CT, Mammo, MRI) and the histopathology caption / closed / open VQA tasks & All four task formats (closed, open, grounding, report) on every modality (16 splits) \\
Initialization & Released Qwen3-VL-8B-Instruct weights & Stage-1 checkpoint \\
Optimizer & AdamW (default $\beta$, weight decay $0.1$) & AdamW (default $\beta$, weight decay $0.1$) \\
Learning rate & $1\times 10^{-5}$, cosine schedule, $0.03$ warm-up ratio & $1\times 10^{-5}$, cosine schedule, $0.03$ warm-up ratio \\
Gradient clipping & $1.0$ & $1.0$ \\
Precision & bfloat16 & bfloat16 \\
Image / volume budget & \texttt{IMAGE\_MAX\_TOKEN\_NUM}=1024; CT and MRI volumes capped at $384\times 384\times 48$ & same \\
Max sequence length & default & $10{,}000$ tokens \\
Selector budget $k$ & 128 & 128 \\
Distributed & DeepSpeed ZeRO-2; per-device batch and accumulation set per launch & DeepSpeed ZeRO-2; per-device batch and accumulation set per launch \\
\bottomrule
\end{tabular}
}
\end{table}

\subsection{System Prompt Templates}\label{supp:system_prompts}

The full stage-aware system-prompt library used by BreastGPT, comprising three personas (Screening, Diagnosis, Treatment, plus a generic clinician fallback) and the deterministic rule that composes a persona with a modality-specific task instruction, is reproduced verbatim from the codebase in Figure~\ref{fig:role_prompts}. The same library is used both during training and at inference time; the persona slot is selected by the record's \texttt{stage} field, and the instruction slot is filled by task and modality as described in the assembly rule.

\section{BreastStage-Bench Evaluation}\label{supp:benchmark_eval}

\subsection{Benchmark Construction}\label{supp:benchmark_construction}

BreastStage-Bench is constructed to evaluate whether a model can preserve stage-aware clinical reasoning across the full breast cancer workflow rather than merely solve isolated image-classification tasks. We use patient-level stratified sampling and keep all task instances derived from the same patient within the same split. This prevents leakage through repeated images, repeated reports, or semantically equivalent QA pairs.

The benchmark construction follows three principles:
\begin{enumerate}
    \item \textbf{No data leakage}: Strict patient-level separation between training and test sets
    \item \textbf{Task diversity}: All 136 task templates are represented with balanced class distributions
    \item \textbf{Clinical realism}: Test cases reflect real-world prevalence and complexity
\end{enumerate}

BreastStage-Bench contains 12{,}182 evaluation records in total, distributed across the four task families and five modalities as follows:
\begin{itemize}
\setlength{\itemsep}{1pt}
    \item Closed-ended VQA (5{,}369 questions): BUS 987, CT 825, Mammography 1{,}828, MRI 908, Histopathology 821.
    \item Open-ended VQA (2{,}833 questions): BUS 987, CT 825, MRI 908, Histopathology 113.
    \item Ground caption (2{,}910 cases): BUS 1{,}000, CT 510, Mammography 1{,}000, MRI 400.
    \item Caption (70 cases): Histopathology slide-level captioning.
    \item Report generation (1{,}000 cases): multiparametric MRI structured reports.
\end{itemize}

\subsection{Evaluation Metrics}\label{supp:eval_metrics}

\paragraph{Closed-ended VQA.}
We adopt accuracy as the primary metric. For multiple-choice questions, we extract model responses using robust regular expressions matching option patterns (A/B/C/D), answer text, and common variants such as ``Option A'' or ``the correct answer is A''. If no valid option is detected in the first 100 tokens of the response, the prediction is marked invalid and counted as incorrect. This avoids rewarding models for verbose but non-committal answers.

\paragraph{Open-ended VQA.}
Unlike LLM-as-judge protocols commonly used in recent multimodal evaluation toolkits~\cite{yu2023mmvetv1,yu2024mmvetv2,duan2024vlmevalkit}, BreastStage-Bench reports automatic semantic and lexical scores for open-ended responses, and uses breast-expert sampling validation to audit the clinical correctness of references and representative outputs. For normalized open-ended scores, each evaluated response receives a score $s_i \in [0, 1]$ according to the task-specific scoring rubric, and the total score is:
\begin{equation}
S = \frac{\sum_{i=1}^N s_i}{N} \times 100\%
\end{equation}

The few-shot prompt includes 9 in-context examples covering fully correct (1.0), entirely incorrect (0.0), and various ``partially correct'' responses (0.25, 0.5, 0.75). The evaluator is instructed to prioritize clinical correctness over surface-form similarity and to penalize unsupported diagnosis, missing contraindications, or recommendations inconsistent with the specified clinical stage.

\paragraph{Grounding Tasks.}
For lesion localization, we compute Intersection over Union (IoU) between predicted and ground-truth bounding boxes:
\begin{equation}
\text{IoU} = \frac{\text{Area}(\text{pred} \cap \text{gt})}{\text{Area}(\text{pred} \cup \text{gt})}
\end{equation}
A prediction is considered correct if IoU $\geq 0.5$. We report mean IoU and accuracy@0.5 across all grounding tasks. Predictions with malformed coordinate syntax are counted as invalid and assigned IoU 0.

\paragraph{Report Generation.}
We evaluate generated reports using:
\begin{itemize}
    \item \textbf{BLEU-4}: Measures n-gram overlap with reference reports
    \item \textbf{BERT Score}: Captures semantic similarity using contextual embeddings (microsoft/BiomedNLP-PubMedBERT-base)
    \item \textbf{ROUGE-1}: Assesses recall of unigrams
\end{itemize}
The weighted composite score is: $\text{Wtd} = 0.5 \times \text{BERT} + 0.25 \times \text{BLEU} + 0.25 \times \text{ROUGE-1}$.



\subsection{Breast Expert Validation}\label{supp:expert_validation}

BreastStage and BreastStage-Bench are constructed with a structured expert sampling audit rather than relying only on automatic filters or LLM-generated text. Three board-certified breast specialists (two breast surgeons and one breast radiologist, each with at least five years of post-residency experience) independently review a stratified subset of BreastStage-Bench, scoring each item along three pre-registered dimensions that mirror the claim made in the main text: \emph{task validity}, \emph{answer correctness}, and \emph{clinical consistency}. The audit covers all five modalities (BUS, mammography, CT, MRI, histopathology) and all task families (closed VQA, open VQA, ground caption, report).

\paragraph{Sampling protocol.}
Sampling is stratified by (modality, task template) so that every one of the 136 task templates contributes evenly: each specialist independently audits 5 random samples per template, yielding 680 per-specialist reviews and a pooled review pool of 2{,}040 \emph{(record, specialist)} judgments. Specialists are blind to which LLM stage produced each record. Each item is judged on the following three binary dimensions:

\begin{itemize}
\item \textbf{Task validity} — the question is well-posed for the indicated clinical stage and the requested evidence is identifiable in the image (or in the source report, for report-only items).
\item \textbf{Answer correctness} — the gold answer is logically consistent with the question and contains no claims unsupported by the image or source report (no hallucinated lesions, biomarkers, or risk categories).
\item \textbf{Clinical consistency} — the assigned clinical stage (screening / diagnosis / treatment) and the modality-specific terminology (BI-RADS category, ACR density, molecular subtype, etc.) match the intended clinical role and current breast-oncology practice.
\end{itemize}

\paragraph{Disagreement resolution.}
Items flagged by at least one but not all specialists are routed to a \emph{consensus review}: the three specialists meet, the contested record is re-examined together, and a single binary decision is recorded by majority vote (with the breast radiologist breaking ties on imaging-specific calls and the senior breast surgeon breaking ties on management calls). Templates whose post-consensus failure rate exceeds 10\,\% are routed back to the responsible LLM prompt for revision, after which the affected records are regenerated and re-audited.

We report per-dimension approval rate and inter-rater agreement (Fleiss's $\kappa$) in \cref{tab:expert_validation}. Across all three dimensions the post-revision approval rate exceeds 95\,\%, and Fleiss's $\kappa$ stays in the substantial-agreement range ($\kappa{\in}[0.74,0.86]$), indicating that the audit signal is reliable rather than dominated by a single specialist's idiosyncrasies.

\begin{table}[H]
\centering
\caption{Expert-validation results on BreastStage-Bench. \textit{Pre-rev.} = approval rate on the first audit pass; \textit{Post-rev.} = rate after the consensus-review and prompt-revision loop. $\kappa$ is Fleiss's $\kappa$ across the three specialists on the binary approval decision. \textit{Reviews} counts \emph{(record, specialist)} judgments; \emph{Answer correctness} is computed on items that carry a gold answer or free-text generation, hence the smaller pool. \textit{Re-gen.} = number of task templates routed to prompt revision.}
\label{tab:expert_validation}
\small
\setlength{\tabcolsep}{4pt}
\begin{tabular}{@{}l r r r c r@{}}
\toprule
\textbf{Dimension} & \textbf{Reviews} & \textbf{Pre-rev.} & \textbf{Post-rev.} & \boldmath$\kappa$ & \textbf{Re-gen.} \\
\midrule
Task validity         & 2{,}040 & 91.4\% & 98.2\% & 0.78 & 14 \\
Answer correctness    & 1{,}620 & 90.6\% & 97.4\% & 0.79 & 11 \\
Clinical consistency  & 2{,}040 & 96.6\% & 99.8\% & 0.86 & \phantom{0}2 \\
\midrule
\textbf{Overall (any-dim. fail)} & 2{,}040 & \textbf{86.1\%} & \textbf{96.3\%} & 0.74 & 22 \\
\bottomrule
\end{tabular}
\end{table}

\paragraph{Per-stage breakdown.}
Post-revision approval is stable across clinical stages (screening 96.5\%, diagnosis 96.1\%, treatment 95.8\%), suggesting the LLM-generated content is not systematically biased toward any single stage. The treatment stage has the lowest pre-revision \emph{answer correctness} score, 86.1\%, because biomarker and prognostic claims are easier for the LLM to over-generate than morphological descriptions; tightening the treatment-stage prompts (constraining biomarker mentions to those present in the source pathology report) raised this dimension to 96.4\% post-revision. The per-stage record counts of the final pipeline export are summarised in \cref{tab:task_taxonomy_detailed} (1{,}068{,}187 screening, 680{,}078 diagnosis, 101{,}585 treatment-stage pairs).

\paragraph{Limitations of this audit.}
Three specialists, while sufficient for $\kappa$ to be meaningful, is below the panel sizes used in formal radiology guideline development; we therefore treat this audit as a quality-assurance pass rather than a clinical gold standard. The full auditing log -- per-record flag, dimension scored, and (where applicable) the resulting prompt revision -- is released alongside the dataset to allow downstream re-auditing.

\subsection{Baseline Model Configuration}\label{supp:baseline_config}

All baseline models are evaluated zero-shot with the same instruction-formatted prompts used for BreastGPT. For proprietary models (GPT-5.4, Claude-opus-4-6, Claude-sonnet-4-6, Gemini-3.1-Flash, Gemini-3.1-Pro), we use the official APIs with temperature 0 and max\_tokens 512. For open-source models we use Hugging Face Transformers with greedy decoding. Input images are preprocessed according to each model's recommended processor; for multi-image or multi-sequence cases all models receive the same ordered visual inputs and the same textual task context.

\section{Additional Results and Ablations}\label{supp:additional_results}

\subsection{Per-Modality Performance Breakdown}\label{supp:per_modality}

Table~\ref{tab:per_modality_breakdown} provides a per-modality performance breakdown for BreastGPT and the strongest competing baselines. This view complements the stage-level tables in the main paper by isolating whether improvements are driven by a single easy modality or are distributed across heterogeneous imaging sources.

\begin{table}[H]
\centering
\caption{Per-modality performance breakdown on BreastStage-Bench closed-ended VQA.}
\label{tab:per_modality_breakdown}
\small
\begin{tabular}{l|ccccc|c}
\toprule
\textbf{Model} & \textbf{BUS} & \textbf{CT} & \textbf{Mammo} & \textbf{MRI} & \textbf{Histo} & \textbf{Avg} \\
\midrule
BreastGPT (cluster) & \textbf{86.81} & \underline{77.21} & \textbf{75.00} & \textbf{82.86} & \textbf{71.38} & \textbf{78.65} \\
GPT-5.4 & 64.89 & \textbf{78.55} & \underline{68.51} & 41.43 & 32.28 & \underline{57.13} \\
Gemini-3.1-Pro & \underline{68.09} & 73.33 & 50.21 & 47.14 & 46.53 & 57.06 \\
Lingshu & 58.94 & \textbf{78.55} & 39.89 & \underline{54.29} & \underline{51.52} & 56.64 \\
\bottomrule
\end{tabular}
\end{table}

\subsection{Extended Baseline Comparisons}
\label{subsec:extended_baselines}

\Cref{tab:vqa_extended,tab:caption_report_extended} report the same VQA and Caption/Report metrics as the main paper's \cref{tab:vqa,tab:caption_report}, but extended to eight additional baseline models (Grok-4.1-Fast, Gemma-4, GLM-4.6V-Flash, LLaVA-OneVision-1.5, HealthGPT, Hulu-Med, MedDr, RadFM) covering both proprietary and recently released open / medical VLMs. None of these baselines beat BreastGPT on any cell, so for compactness the main paper limits its comparison to the original cohort; this section confirms that the conclusion holds against a broader set.

\begin{table}[H]
\caption{VQA performance (\%) on BreastStage-Bench for eight additional baselines (extends \cref{tab:vqa}). Same column structure as the main table; refer to it for column legend.}
\label{tab:vqa_extended}
\centering
\scriptsize
\setlength{\tabcolsep}{2pt}
\resizebox{\textwidth}{!}{%
\begin{tabular}{l c *{9}{c} c | *{7}{c} c}
\toprule
& & \multicolumn{10}{c|}{\textbf{Closed-ended VQA (Accuracy, \%)}} & \multicolumn{8}{c}{\textbf{Open-ended VQA (normalized Score, \%)}} \\
\cmidrule(lr){3-12}\cmidrule(lr){13-20}
& & \multicolumn{4}{c}{Screening} & \multicolumn{3}{c}{Diagnosis} & \multicolumn{2}{c}{Treatment} & & \multicolumn{3}{c}{Screening} & \multicolumn{2}{c}{Diagnosis} & \multicolumn{2}{c}{Treatment} & \\
\cmidrule(lr){3-6}\cmidrule(lr){7-9}\cmidrule(lr){10-11}\cmidrule(lr){13-15}\cmidrule(lr){16-17}\cmidrule(lr){18-19}
Model & \#P & BUS & CT & Mam & MRI & BUS & Mam & MRI & MRI & His & Avg & BUS & CT & MRI & BUS & MRI & MRI & His & Avg \\
\midrule
\rowcolor{gray!10}\multicolumn{20}{l}{\textit{Proprietary Models}} \\
Grok-4.1-Fast & -- & 57.23 & 76.61 & 30.74 & 40.95 & \textbf{49.47} & \textbf{22.82} & 44.23 & 31.75 & 36.42 & 43.36 & 44.76 & 43.43 & 43.95 & 43.06 & 43.94 & 41.74 & 40.50 & 43.05 \\
\rowcolor{gray!10}\multicolumn{20}{l}{\textit{Open-Source Models}} \\
Gemma-4 & 8B & \textbf{62.98} & \textbf{78.18} & \underline{44.57} & 38.57 & 47.87 & 18.17 & 48.43 & 39.68 & 40.68 & 46.57 & 43.29 & 44.20 & 43.46 & 42.47 & 44.49 & 42.40 & 40.12 & 42.92 \\
GLM-4.6V-Flash & 9B & \underline{58.94} & 73.09 & 37.34 & 41.90 & 45.48 & 12.16 & 48.60 & \textbf{53.97} & \underline{49.70} & \underline{46.80} & 42.69 & 43.70 & 42.78 & 42.05 & 43.54 & 42.47 & 40.28 & 42.50 \\
LLaVA-OneVision-1.5 & 8B & 58.09 & 68.85 & 30.32 & 40.95 & 41.76 & 12.31 & \textbf{50.17} & \textbf{53.97} & \textbf{50.18} & 45.18 & -- & 56.06 & 59.84 & -- & 59.19 & 58.12 & 45.25 & 55.69 \\
\rowcolor{gray!10}\multicolumn{20}{l}{\textit{Medical-Specific Models}} \\
HealthGPT & 32B & 47.02 & 24.48 & 41.38 & \underline{45.71} & 34.04 & 21.17 & 43.01 & 42.86 & 20.22 & 35.54 & 54.66 & 50.17 & 53.08 & 51.92 & 59.38 & 55.27 & 46.41 & 52.98 \\
Hulu-Med & 7B & \textbf{62.98} & \underline{77.09} & \textbf{48.19} & 31.90 & \underline{49.20} & 15.62 & 41.43 & \underline{50.79} & \textbf{50.18} & \textbf{47.49} & \underline{58.98} & 60.84 & \textbf{75.46} & \underline{55.95} & \underline{72.28} & 59.48 & 50.18 & \underline{61.88} \\
MedDr & 40B & 48.51 & 47.39 & 25.21 & \textbf{52.38} & 35.37 & \underline{22.52} & \underline{49.83} & 43.65 & 39.46 & 40.48 & \textbf{61.46} & \textbf{64.52} & \underline{67.68} & \textbf{62.83} & \textbf{73.03} & \underline{64.86} & \textbf{50.46} & \textbf{63.55} \\
RadFM & 14B & 26.81 & 39.39 & 16.38 & 36.67 & 23.94 & 5.86 & 17.13 & 11.90 & 4.63 & 20.30 & 50.51 & \underline{61.07} & 67.63 & 50.80 & 69.15 & \textbf{66.76} & \underline{50.44} & 59.48 \\
\bottomrule
\end{tabular}%
}
\end{table}

\begin{table}[H]
\caption{Caption and Report Generation Performance (\%) for eight additional baselines (extends \cref{tab:caption_report}). Same column structure as the main table.}
\label{tab:caption_report_extended}
\centering
\small
\resizebox{\textwidth}{!}{%
\begin{tabular}{p{3.5cm} c
  *{2}{>{\centering\arraybackslash}p{1cm}}
  >{\centering\arraybackslash}p{1cm}
  *{2}{>{\centering\arraybackslash}p{1cm}}
  >{\centering\arraybackslash}p{1cm}
  >{\centering\arraybackslash}p{1cm}|
  >{\centering\arraybackslash}p{1cm}}
\toprule
& & \multicolumn{7}{c|}{\textbf{Caption}} & \textbf{Report} \\
\cmidrule(lr){3-9}\cmidrule(lr){10-10}
& & \multicolumn{2}{c}{\textbf{BUS}} & \textbf{CT} & \multicolumn{2}{c}{\textbf{Mammo}} & \textbf{Histo} & \textbf{MRI} & \textbf{MRI} \\
\cmidrule(lr){3-4}\cmidrule(lr){5-5}\cmidrule(lr){6-7}\cmidrule(lr){8-8}\cmidrule(lr){9-9}\cmidrule(lr){10-10}
Model & \#P & IoU & Wtd & Wtd & IoU & Wtd & Wtd & Wtd & Wtd \\
\hline
\rowcolor{gray!10}\multicolumn{10}{l}{\textit{Proprietary Models}} \\
\hline
Grok-4.1-Fast & -- & 13.40 & 47.30 & 45.12 & 2.52 & 48.09 & 45.15 & \textbf{49.02} & \underline{50.47} \\
\hline
\rowcolor{gray!10}\multicolumn{10}{l}{\textit{Open-Source Models}} \\
\hline
Gemma-4 & 8B & \textbf{37.82} & 46.34 & 44.45 & \textbf{8.01} & 46.33 & 44.04 & 47.87 & 49.16 \\
GLM-4.6V-Flash & 9B & \underline{28.03} & 43.01 & 43.43 & 6.01 & 43.15 & 43.53 & 44.26 & 48.21 \\
LLaVA-OneVision-1.5 & 8B & 14.99 & \underline{47.68} & \underline{49.04} & 4.53 & \underline{48.41} & \underline{46.91} & \underline{48.62} & \textbf{50.62} \\
\hline
\rowcolor{gray!10}\multicolumn{10}{l}{\textit{Medical-Specific Models}} \\
\hline
HealthGPT & 32B & 22.56 & 43.11 & 40.05 & \underline{7.69} & 44.40 & 46.63 & 46.93 & 48.18 \\
Hulu-Med & 7B & 5.75 & \textbf{48.67} & 48.40 & 0.41 & \textbf{48.74} & \textbf{47.25} & 48.49 & 50.33 \\
MedDr & 40B & 0.17 & \underline{47.68} & \textbf{50.25} & 0.38 & 48.34 & 44.73 & 47.04 & 46.60 \\
RadFM & 14B & - & 41.90 & 38.13 & - & 43.42 & 41.23 & 46.53 & 43.27 \\
\bottomrule
\end{tabular}%
}
\end{table}

\subsection{Open-ended VQA Raw Metric Breakdown}
\label{subsec:open_vqa_raw}

\Cref{tab:open_vqa_raw} reports the three constituent metrics (BERTScore F1, BLEU, ROUGE-1) underlying the weighted score reported in \cref{tab:vqa}. Closed-ended VQA is omitted because its only metric is accuracy.

\begin{table}[H]
\centering
\caption{Open-ended VQA raw metrics per cell. Each entry is BERTScore F1 / BLEU / ROUGE-1 (\%). BreastGPT in cyan.}
\label{tab:open_vqa_raw}
\scriptsize
\setlength{\tabcolsep}{2pt}
\resizebox{\textwidth}{!}{%
\begin{tabular}{l c *{7}{c}}
\toprule
& & \multicolumn{3}{c}{Screening} & \multicolumn{2}{c}{Diagnosis} & \multicolumn{2}{c}{Treatment} \\
\cmidrule(lr){3-5}\cmidrule(lr){6-7}\cmidrule(lr){8-9}
Model & \#P & BUS & CT & MRI & BUS & MRI & MRI & His \\
\midrule
\rowcolor{gray!10}\multicolumn{9}{l}{\textit{Proprietary Models}} \\
GPT-5.4 & -- & 88.88/1.62/34.33 & 86.50/1.65/22.73 & 91.82/11.65/44.10 & 87.25/1.45/25.71 & 90.82/14.09/41.05 & 90.57/10.10/41.04 & 83.67/0.57/11.02 \\
Claude-opus-4-6 & -- & 82.55/0.45/7.73 & 81.86/0.66/6.70 & 82.75/1.25/8.24 & 82.05/0.30/6.81 & 82.95/1.87/9.90 & 81.76/0.73/7.40 & 79.96/0.17/3.27 \\
Claude-sonnet-4-6 & -- & 82.54/0.46/8.97 & 81.71/0.88/8.16 & 82.71/1.26/10.56 & 81.85/0.34/7.94 & 82.83/1.82/12.46 & 80.90/0.51/8.15 & 79.90/0.14/4.32 \\
Gemini-3.1-Flash & -- & 86.50/1.51/20.51 & 86.04/2.68/18.08 & 85.60/2.51/15.81 & 84.66/0.68/13.19 & 85.36/3.26/16.15 & 83.07/1.12/9.34 & 82.27/0.34/6.37 \\
Grok-4.1-Fast & -- & 83.70/0.56/11.10 & 82.32/0.82/8.25 & 82.83/0.90/9.24 & 82.21/0.30/7.51 & 82.33/1.29/9.83 & 80.84/0.34/4.93 & 79.63/0.13/2.60 \\
\rowcolor{gray!10}\multicolumn{9}{l}{\textit{Open-Source Models}} \\
Qwen2.5-VL & 3B & 87.99/2.12/19.97 & 86.85/2.98/16.89 & 87.59/3.41/18.86 & 86.69/1.31/14.75 & 87.48/5.50/21.70 & 85.49/2.08/15.08 & 84.46/0.49/9.80 \\
Qwen2.5-VL & 7B & 86.11/1.13/11.45 & 85.82/2.08/13.72 & 87.36/2.70/20.45 & 85.06/0.48/9.36 & 86.98/4.61/20.11 & 83.92/1.32/9.62 & 83.34/0.32/8.94 \\
Qwen3-VL & 4B & 84.69/0.98/10.42 & 89.65/5.08/37.59 & 86.29/2.10/16.16 & 84.23/0.91/10.36 & 87.78/4.95/25.31 & 85.61/1.86/22.86 & 80.65/0.22/3.52 \\
Qwen3-VL & 8B & 84.40/1.00/9.98 & 86.40/2.64/18.68 & 83.89/1.73/9.90 & 83.76/0.74/9.11 & 84.43/2.78/13.30 & 82.52/1.14/10.41 & 80.51/0.23/3.60 \\
MiMo-VL & 7B & 82.27/0.52/4.90 & 88.14/4.29/43.35 & 89.97/19.76/55.10 & 81.32/0.37/3.75 & 90.36/22.57/59.33 & 89.91/16.07/56.42 & 78.85/0.14/1.59 \\
InternVL3.5 & 8B & 87.86/4.39/22.84 & 89.02/14.39/39.77 & 90.19/7.99/37.85 & 86.56/2.69/19.37 & 90.65/12.68/41.59 & 89.44/7.36/37.54 & 84.86/0.50/15.80 \\
Gemma-4 & 8B & 82.79/0.52/7.05 & 83.20/1.12/9.26 & 82.71/1.05/7.39 & 81.86/0.32/5.83 & 83.17/1.91/9.71 & 81.59/0.68/5.75 & 79.20/0.05/2.04 \\
GLM-4.6V-Flash & 9B & 82.16/0.71/5.73 & 83.03/1.35/7.38 & 82.38/1.16/5.20 & 81.52/0.57/4.58 & 82.90/1.69/6.65 & 82.17/0.89/4.65 & 79.42/0.19/2.10 \\
LLaVA-OneVision-1.5 & 8B & -- & 90.14/6.57/37.39 & 91.82/11.65/44.09 & -- & 90.82/14.09/41.03 & 90.57/10.10/41.24 & 84.88/0.39/10.84 \\
\rowcolor{gray!10}\multicolumn{9}{l}{\textit{Medical-Specific Models}} \\
Lingshu & 7B & 89.47/3.66/27.78 & 87.79/4.01/19.76 & 89.29/5.04/30.02 & 87.59/1.80/18.81 & 88.65/8.22/29.52 & 86.70/3.17/21.84 & 83.57/0.30/7.32 \\
HuatuoGPT-V & 7B & 88.99/3.60/25.86 & 88.43/5.17/23.85 & 90.20/6.84/35.48 & 87.66/2.25/19.23 & 89.36/10.04/31.58 & 88.17/5.16/30.22 & 85.07/0.88/12.08 \\
HealthGPT & 32B & 89.69/5.73/33.52 & 86.17/7.50/20.84 & 87.02/10.10/28.18 & 88.90/3.75/26.11 & 89.91/16.43/41.27 & 87.84/10.10/35.30 & 84.83/0.71/15.28 \\
Hulu-Med & 7B & 91.14/9.89/43.73 & 90.99/14.81/46.56 & 95.29/40.64/70.61 & 90.02/6.34/37.42 & 93.60/37.93/64.00 & 90.64/10.31/46.35 & 87.35/3.33/22.70 \\
MedDr & 40B & 91.22/20.02/43.39 & 90.46/31.04/46.13 & 92.69/26.61/58.74 & 91.85/19.38/48.23 & 93.70/40.07/64.63 & 91.82/21.43/54.38 & 87.12/3.25/24.35 \\
RadFM & 14B & 87.50/3.86/23.19 & 89.66/24.94/40.04 & 93.08/25.80/58.57 & 86.53/9.71/20.44 & 92.94/30.76/59.96 & 91.86/26.74/56.57 & 87.00/3.21/24.55 \\
\rowcolor{cyan!15}\textbf{Qwen3-VL (SFT)} & \textbf{8B} & \textbf{98.85/87.18/92.83} & \textbf{99.04/89.22/94.24} & \textbf{99.11/85.25/94.82} & \textbf{98.26/81.90/88.38} & \textbf{99.03/87.90/95.17} & \textbf{97.65/72.02/86.10} & \textbf{90.73/11.46/38.93} \\
\rowcolor{cyan!15}\textbf{BreastGPT (cluster)} & \textbf{8B} & \textbf{99.15/90.67/94.91} & \textbf{99.07/88.56/94.45} & \textbf{99.27/87.53/95.87} & \textbf{98.63/85.08/90.60} & \textbf{99.12/88.94/95.68} & \textbf{98.09/75.47/88.08} & \textbf{92.69/18.88/50.96} \\
\bottomrule
\end{tabular}%
}
\end{table}

\subsection{3D Grounding IoU on CT and MRI}
\label{subsec:caption_3d_iou}

None of the 21 baseline models in our evaluation cohort can localise lesions in 3D volumetric modalities (CT, MRI). Their bbox outputs are uniformly 2-D coordinate quadruples, which are dimensionally incompatible with the 6-element 3-D ground-truth boxes; consequently, on every GT-positive 3-D sample they fail to overlap the ground truth at all. The grounding-IoU numbers a few baselines report on these modalities are entirely a by-product of true-negative credit on the no-abnormality samples, not of any spatial localisation. We therefore report 3-D grounding IoU only for BreastGPT and its controlled SFT counterpart in \cref{tab:caption_3d_iou}.

\begin{table}[H]
\centering
\caption{Grounding IoU (\%) on the 3D modalities (CT, MRI), restricted to the two models that emit volumetric (6-D) bounding boxes. IoU credits true negatives as 1.0 and penalises false positives as 0. All 21 other baselines produce 2-D outputs that are dimensionally incompatible with the 3-D ground truth and are omitted here.}
\label{tab:caption_3d_iou}
\small
\setlength{\tabcolsep}{4pt}
\begin{tabular}{l c c c}
\toprule
Model & \#P & \textbf{CT IoU} & \textbf{MRI IoU} \\
\midrule
\rowcolor{cyan!15}\textbf{Qwen3-VL (SFT)} & \textbf{8B} & \underline{3.89} & \underline{11.98} \\
\rowcolor{cyan!15}\textbf{BreastGPT (cluster)} & \textbf{8B} & \textbf{5.12} & \textbf{33.49} \\
\bottomrule
\end{tabular}
\end{table}

\subsection{BreastGPT Grounding Recognition Breakdown}
\label{subsec:breastgpt_iou}

\Cref{tab:breastgpt_iou_breakdown} decomposes the IoU column of \cref{tab:caption_report} for BreastGPT (cluster) into its true-negative and false-positive components: TN counts items with no GT bbox where the model also abstains (correct ``no abnormality''), FP counts items where the model hallucinates a bbox on a normal case, and \textit{Pred-with-GT} counts items where BreastGPT attempted a bbox on a GT-positive case.

\begin{table}[H]
\centering
\caption{BreastGPT (cluster) grounding-IoU decomposition. IoU credits true negatives (TN) as 1.0 and penalises false positives (FP) as 0. \textit{Pred-with-GT} counts items where BreastGPT attempted a bbox on a GT-positive case; $N$ is the total number of items.}
\label{tab:breastgpt_iou_breakdown}
\small
\begin{tabular}{l r r r r r}
\toprule
\textbf{Modality} & \textbf{IoU (\%)} & \textbf{TN} & \textbf{FP} & \textbf{Pred-with-GT} & \textbf{N} \\
\midrule
BUS & 79.59 & 0 & 0 & 1000 & 1000 \\
CT & 5.12 & 33 & 70 & 333 & 510 \\
Mammo & 23.14 & 0 & 2 & 998 & 1000 \\
MRI & 33.49 & 108 & 60 & 202 & 400 \\
\bottomrule
\end{tabular}
\end{table}

\subsection{Inference Efficiency Analysis}
\label{subsec:efficiency}

We profile the actual cost of running BreastGPT on the histopathology WSI subset of BreastStage-Bench at different token budgets $k$, including a no-limit baseline that disables the selector and feeds every patch token of the slide into the LLM. Protocol: bf16, batch size 1, 5 warm-up + 30 timed forward passes, on a single GPU. Latency is measured with \texttt{torch.cuda.Event} and \texttt{torch.cuda.synchronize()} so that timing reflects the actual GPU execution rather than asynchronous launches. Selector latency is the wall-clock time inside the greedy coverage routine; prefill latency is the first-token forward through the LLM after selection; total latency is selector + prefill.

\begin{table}[H]
\centering
\caption{Inference efficiency of BreastGPT (cluster) on a representative histopathology WSI (5{,}987 patches before selection). ``LLM in.\ tok.''\ is the total LLM input length after selection (visual $k$ tokens plus prompt). The ``no limit'' row sets $k$ above the patch count, effectively disabling the selector and feeding every patch token to the LLM.}
\label{tab:inference_efficiency}
\small
\setlength{\tabcolsep}{4pt}
\begin{tabular}{c|c|c|c|c|c|c}
\toprule
\textbf{$k$} & \textbf{LLM in.\ tok.} & \textbf{Prefill (ms)} & \textbf{Selector (ms)} & \textbf{Decode (ms/tok)} & \textbf{Total (ms)} & \textbf{Peak Mem (GB)} \\
\midrule
1     & 145   & 57.9   & 0.4    & 42.6  & 58.2    & 16.96 \\
8     & 152   & 67.3   & 0.9    & 45.9  & 68.2    & 16.96 \\
16    & 160   & 78.1   & 1.5    & 50.4  & 79.6    & 16.96 \\
32    & 176   & 89.4   & 2.6    & 48.0  & 91.9    & 16.96 \\
64    & 208   & 123.1  & 4.8    & 54.5  & 127.9   & 16.96 \\
\rowcolor{cyan!10}
128   & 272   & 191.4  & 9.3    & 68.2  & 200.6   & 16.97 \\
256   & 400   & 328.4  & 18.2   & 95.5  & 346.6   & 16.97 \\
512   & 656   & 605.7  & 36.1   & 150.7 & 641.8   & 16.98 \\
no limit & 5{,}144 & 5{,}512.6 & --- & 1{,}129.5 & 6{,}545.7 & 17.95 \\
\bottomrule
\end{tabular}
\end{table}

Three observations follow from these numbers; the per-stage latency breakdown is plotted in main-paper \cref{fig:latency}.
\emph{(i) Prefill and decode scale roughly linearly with $k$.} From $k{=}1$ to $k{=}512$ the LLM input length grows from 145 to 656 tokens, and prefill grows in step from 57.9 to 605.7 ms; decode time per output token grows from 42.6 to 150.7 ms, since the KV-cache that each new token must attend to also lengthens with $k$. The chosen $k{=}128$ costs 191 ms of prefill, $3.2\times$ less than $k{=}512$, at $>$99\% of its task quality.
\emph{(ii) Without selection, the LLM forward dominates the entire pipeline.} The no-limit baseline raises the LLM input to 5{,}144 tokens; prefill jumps to 5.5 s and per-token decode to 1.1 s, for a total of 6.5 s per question—about $33\times$ slower than $k{=}128$ on the same slide. The selector itself only takes 1.0 s in this regime, so the cost saved is overwhelmingly LLM-side, exactly as expected for a method that compresses the LLM input.
\emph{(iii) Memory grows slowly under selection and jumps only at no limit.} Peak GPU memory rises from 16.96 GB at $k{=}1$ to 16.98 GB at $k{=}512$ and then to 17.95 GB without selection, a $\sim$1 GB jump that scales with the LLM input length. BreastGPT therefore runs on a single 24 GB GPU at any selected budget; the no-limit configuration would exceed that envelope on larger slides.

\section{Detailed Ablation Results}\label{supp:detailed_ablation}

\subsection{Numerical Results for WSI Branch Ablation}\label{supp:wsi_ablation_full}

Table~\ref{tab:wsi_ablation_numerical} reports an internal design ablation of the GigaPixel branch, complementary to the Qwen3-VL (SFT, 32-patch WSI) baseline reported in the main paper. This ablation isolates the contribution of each subcomponent inside the GigaPixel branch itself: average pooling loses spatial heterogeneity, truncation ignores sparse diagnostic regions, V-V-only selection may miss query-relevant evidence, and T-V-only selection may overfocus on a small salient region.

\begin{table}[H]
\centering
\caption{Numerical results for WSI branch ablation study on histopathology tasks.}
\label{tab:wsi_ablation_numerical}
\small
\resizebox{\textwidth}{!}{%
\begin{tabular}{l|c|c|c|c}
\toprule
\textbf{Configuration} & \textbf{Closed Acc (\%)} & \textbf{Caption BERT} & \textbf{Caption BLEU} & \textbf{Caption R-1} \\
\midrule
CONCH + LongNet + V-V only & 69.4 & 0.86 & 0.033 & 0.248 \\
CONCH + LongNet + T-V + V-V (full) & \textbf{71.4} & \textbf{0.88} & \textbf{0.038} & \textbf{0.267} \\
\bottomrule
\end{tabular}%
}
\end{table}

The results demonstrate that each component contributes incrementally. LongNet contextualization improves closed-ended accuracy by 9.5 points over average pooling, showing that slide-level context is essential for WSI reasoning. Adding V-V coverage gains another 6.6 points, indicating that global representativeness matters beyond sequential context aggregation. Incorporating T-V coverage yields the final 3.8-point improvement, confirming that clinical-query alignment is necessary once the selector is asked to compress the slide into a very small token budget.

\subsection{Numerical Results for Visual Token Budget Sweep}
\label{subsec:token_budget_full}

Tables~\ref{tab:token_budget_vqa_full} and \ref{tab:token_budget_caption_full} provide the complete numerical breakdown of the visual token budget sweep summarized in Figure~\ref{fig:token_budget}. We sweep $k \in \{1, 8, 16, 32, 64, 128, 256, 512\}$ and report performance on every modality and every task type in BreastStage-Bench: VQA (closed- and open-ended) across screening, diagnosis, and treatment stages, as well as caption and report generation. This is intentionally exhaustive so that the operating-point choice ($k{=}128$) used throughout the main paper can be audited per-modality, not only on histopathology.

\begin{table}[H]
\centering
\caption{Token budget sweep on the VQA task across screening, diagnosis, and treatment stages, under closed-ended and open-ended settings. ``Histo.'' denotes histopathology.}
\label{tab:token_budget_vqa_full}
\setlength{\tabcolsep}{3pt}
\renewcommand{\arraystretch}{1.05}
\resizebox{\textwidth}{!}{%
\begin{tabular}{c | cccc | ccc | ccc | cc | cc | cc}
\toprule
 & \multicolumn{7}{c|}{\textbf{Screen}}
 & \multicolumn{5}{c|}{\textbf{Diagnosis}}
 & \multicolumn{4}{c}{\textbf{Treatment}} \\
\cmidrule(lr){2-8}\cmidrule(lr){9-13}\cmidrule(lr){14-17}
\textbf{Token}
 & \multicolumn{4}{c|}{closed}
 & \multicolumn{3}{c|}{open}
 & \multicolumn{3}{c|}{closed}
 & \multicolumn{2}{c|}{open}
 & \multicolumn{2}{c|}{closed}
 & \multicolumn{2}{c}{open} \\
\cmidrule(lr){2-5}\cmidrule(lr){6-8}\cmidrule(lr){9-11}\cmidrule(lr){12-13}\cmidrule(lr){14-15}\cmidrule(lr){16-17}
 & BUS & CT & Mammo & MRI
 & BUS & CT & MRI
 & BUS & Mammo & MRI
 & BUS & MRI
 & MRI & Histo.
 & MRI & Histo. \\
\midrule
1   & 60.21 & 78.79 & 43.41 & 72.38 & 91.84 & 95.66 & 94.80 & 51.06 & 48.80 & 65.39 & 87.82 & 94.63 & 40.48 & 66.63 & 89.63 & 56.82 \\
8   & 75.32 & 78.43 & 60.43 & 80.95 & 93.76 & 94.89 & 95.47 & 69.94 & 64.41 & 70.28 & 90.53 & 94.99 & 46.83 & 71.25 & 89.22 & 63.78 \\
16  & 84.04 & 77.70 & 72.88 & 81.43 & 95.41 & 94.83 & 95.83 & 75.26 & 67.42 & 78.32 & 92.02 & 95.26 & 59.52 & 71.01 & 89.67 & 66.12 \\
32  & 86.17 & 77.70 & 74.05 & 85.24 & 95.80 & 95.03 & 95.58 & 76.86 & 68.92 & 81.29 & 93.08 & 95.71 & 62.70 & 71.01 & 89.96 & 66.53 \\
64  & 86.81 & 76.73 & 74.69 & 83.81 & 95.96 & 95.48 & 95.58 & 77.66 & 68.77 & 81.12 & 92.94 & 95.72 & 61.11 & 71.13 & 89.95 & 66.08 \\
128 & 86.81 & 77.21 & 75.00 & 82.86 & 95.97 & 95.29 & 95.48 & 77.13 & 68.32 & 81.12 & 93.24 & 95.72 & 61.11 & 71.38 & 89.93 & 63.80 \\
256 & 86.81 & 76.73 & 74.05 & 82.38 & 95.97 & 95.51 & 95.75 & 77.13 & 68.02 & 81.29 & 93.24 & 95.61 & 59.52 & 71.01 & 89.97 & 66.05 \\
512 & 86.81 & 77.58 & 74.47 & 81.90 & 95.96 & 95.70 & 95.65 & 77.13 & 68.32 & 82.69 & 93.23 & 95.67 & 57.14 & 71.50 & 89.89 & 64.58 \\
\bottomrule
\end{tabular}%
}
\end{table}

\begin{table}[H]
\centering
\caption{Token budget sweep on the caption and report generation tasks. ``Wtd'' denotes the weighted average of BERT-F1, BLEU, and ROUGE-1. The ``IoU'' column reports the \emph{Ground Caption} task (lesion-grounded captioning), which is only defined for the modalities with bounding-region annotations.}
\label{tab:token_budget_caption_full}
\setlength{\tabcolsep}{2pt}
\renewcommand{\arraystretch}{1.05}
\resizebox{\textwidth}{!}{%
\begin{tabular}{c | ccccc | ccccc | ccccc | cccc | ccccc}
\toprule
 & \multicolumn{15}{c|}{\textbf{Caption}}
 & \multicolumn{9}{c}{\textbf{Report}} \\
\cmidrule(lr){2-16}\cmidrule(lr){17-25}
\textbf{Token}
 & \multicolumn{5}{c|}{BUS}
 & \multicolumn{5}{c|}{CT}
 & \multicolumn{5}{c|}{Mammo}
 & \multicolumn{4}{c|}{Histopathology}
 & \multicolumn{5}{c}{MRI} \\
\cmidrule(lr){2-6}\cmidrule(lr){7-11}\cmidrule(lr){12-16}\cmidrule(lr){17-20}\cmidrule(lr){21-25}
 & BERT-F1 & BLEU & R-1 & Wtd & IoU
 & BERT-F1 & BLEU & R-1 & Wtd & IoU
 & BERT-F1 & BLEU & R-1 & Wtd & IoU
 & BERT-F1 & BLEU & R-1 & Wtd
 & BERT-F1 & BLEU & R-1 & Wtd & IoU \\
\midrule
1   & 92.77 & 36.61 & 61.89 & 71.01 & 1.50  & 93.90 & 38.01 & 65.46 & 72.83 & 4.90  & 93.77 & 38.35 & 64.75 & 72.66 & 0.85  & 83.86 & 18.23 & 32.70 & 54.67 & 89.85 & 26.01 & 54.91 & 65.15 & 24.52 \\
8   & 94.08 & 45.79 & 69.41 & 75.84 & 20.00 & 93.70 & 37.27 & 64.40 & 72.28 & 4.91  & 94.20 & 44.52 & 67.53 & 75.11 & 2.13  & 88.34 & 40.47 & 53.35 & 67.63 & 90.09 & 27.27 & 55.43 & 65.71 & 24.72 \\
16  & 94.63 & 50.24 & 72.44 & 77.99 & 40.88 & 93.77 & 38.18 & 64.69 & 72.61 & 4.93  & 94.67 & 49.48 & 69.88 & 77.18 & 5.00  & 88.04 & 39.57 & 50.99 & 66.66 & 90.37 & 28.82 & 56.52 & 66.51 & 25.41 \\
32  & 94.90 & 52.21 & 73.85 & 78.97 & 63.04 & 93.88 & 38.92 & 65.34 & 73.01 & 4.93  & 94.72 & 50.04 & 70.03 & 77.38 & 12.52 & 87.82 & 37.51 & 49.98 & 65.79 & 90.50 & 30.07 & 57.47 & 67.13 & 26.17 \\
64  & 94.94 & 52.71 & 74.23 & 79.21 & 76.76 & 93.89 & 40.89 & 65.89 & 73.65 & 5.03  & 94.73 & 50.35 & 70.05 & 77.46 & 18.37 & 88.15 & 39.50 & 51.81 & 66.91 & 90.58 & 30.90 & 58.11 & 67.53 & 28.55 \\
128 & 94.97 & 52.91 & 74.42 & 79.32 & 79.59 & 93.76 & 39.81 & 65.30 & 73.16 & 5.12  & 94.75 & 50.74 & 70.31 & 77.64 & 23.14 & 88.12 & 39.61 & 51.25 & 66.78 & 90.61 & 31.24 & 58.24 & 67.67 & 33.49 \\
256 & 95.00 & 53.48 & 74.55 & 79.51 & 80.33 & 93.73 & 39.41 & 65.24 & 73.03 & 5.03  & 94.84 & 51.57 & 70.91 & 78.04 & 23.06 & 88.00 & 37.67 & 50.14 & 65.96 & 90.59 & 31.16 & 58.09 & 67.60 & 29.64 \\
512 & 94.98 & 53.10 & 74.45 & 79.38 & 80.36 & 93.79 & 39.96 & 65.61 & 73.29 & 5.02  & 94.80 & 51.11 & 70.69 & 77.85 & 23.10 & 87.58 & 34.94 & 47.98 & 64.53 & 90.62 & 31.21 & 58.29 & 67.68 & 29.52 \\
\bottomrule
\end{tabular}%
}
\end{table}

Two patterns hold consistently across modalities. \emph{(i) Performance saturates at $k{=}128$.} On VQA closed-ended screening, BUS reaches its plateau at $k{=}64$ (86.81\%) and stays flat through $k{=}512$; CT, Mammo, and MRI similarly fluctuate within $\pm 1$ point past $k{=}64$. On caption generation, the weighted score on BUS rises from 71.01 at $k{=}1$ to 79.32 at $k{=}128$ and gains only 0.06 points by $k{=}512$; Mammo gains only 0.21 points from 128 to 512. On open-ended VQA, all six (modality, stage) cells are within 0.5 points of their maximum by $k{=}128$. \emph{(ii) The ground caption task needs a minimum number of tokens.} BUS ground caption IoU jumps from 1.50 at $k{=}1$ to 79.59 at $k{=}128$ before plateauing, and Mammo ground caption IoU from 0.85 to 23.14. This indicates that a single coverage-selected token is insufficient to anchor a lesion bounding region, but $k{=}128$ already recovers most of the IoU achievable at $k{=}512$. The 128-token budget therefore gives BreastGPT the same downstream inference cost on radiology and pathology while staying on the saturation plateau of every metric we measure.

\section{Qualitative Analysis and Case Studies}\label{supp:qualitative}

\subsection{Case Analysis}\label{supp:case_analysis}

To complement the aggregate metrics in \cref{tab:vqa,tab:caption_report}, we walk through six representative cases drawn from BreastStage-Bench that span all five modalities and all four task families (closed-ended VQA, open-ended VQA, lesion-grounded captioning, and structured report generation). For each case we show the role-conditioned system prompt that BreastGPT receives at inference time (abbreviated to its lead clause), the user instruction, and the reference output that defines the gold answer for the metrics in the main paper. Together they illustrate the breadth of input formats and reasoning styles a single BreastGPT checkpoint must support.

\begin{figure}[H]
\centering
\includegraphics[width=0.32\linewidth]{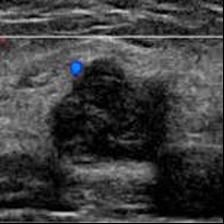}
\begin{promptbox}[{{Case 1 -- BUS, Screening-stage Closed VQA (vascularity attribute)}}]{teal}
\textbf{System.} You are an expert screening breast radiologist. Your clinical focus is on early detection and risk triage across broad populations. You possess a high sensitivity for identifying subtle abnormalities within varied anatomical backgrounds to determine if further investigation is warranted. Now, apply your expertise to the following task: Answer the multiple-choice question based on the provided breast ultrasound image(s). Respond ONLY with the exact option chosen.

\textbf{User.} \texttt{<image>}\\
Classify the blood flow status: No flow, Mild flow, or Abundant flow?\\
Options: (A) No flow \quad (B) Abundant flow \quad (C) Mild flow\\
Answer:

\textbf{Reference output.} (C) Mild flow

\textbf{Comment.} The screening persona drives the model toward a categorical Doppler readout rather than a free-text discussion, and the answer must come from the schema enum so it can be scored by exact-string accuracy.
\end{promptbox}
\caption{Case 1: closed-ended vascularity attribute on a single BUS B-mode frame from BUS-CoT.}
\label{fig:case1}
\end{figure}

\begin{figure}[H]
\centering
\includegraphics[width=0.32\linewidth]{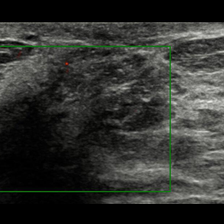}
\begin{promptbox}[{{Case 2 -- BUS, Screening-stage Open-ended VQA (Doppler description)}}]{teal}
\textbf{System.} You are an expert screening breast radiologist. Your clinical focus is on early detection and risk triage across broad populations. You possess a high sensitivity for identifying subtle abnormalities within varied anatomical backgrounds to determine if further investigation is warranted. Now, apply your expertise to the following task: Answer the open-ended question based on the provided breast ultrasound image(s).

\textbf{User.} \texttt{<image>}\\
Describe the degree of vascularity observed within the lesion on Doppler ultrasound.

\textbf{Reference output.} Mild flow is detected within the lesion.

\textbf{Comment.} The same underlying clinical fact as Case 1 is rewritten into a fluent description; this pair is what allows the weighted (BERTScore F1, BLEU, ROUGE-1) score to evaluate semantic equivalence rather than exact-string match.
\end{promptbox}
\caption{Case 2: open-ended Doppler description on a BUS Doppler frame from BUS-CoT.}
\label{fig:case2}
\end{figure}

\begin{figure}[H]
\centering
\includegraphics[width=0.40\linewidth]{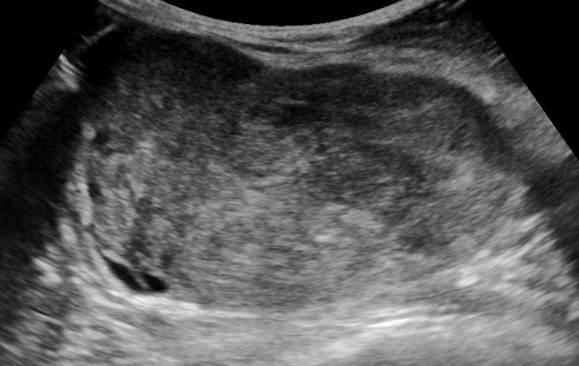}
\begin{promptbox}[{{Case 3 -- BUS, Lesion-grounded Captioning}}]{orange}
\textbf{System.} You are an expert breast clinician. Your clinical focus is on comprehensive breast disease assessment and patient-centered clinical management. You integrate clinical presentation, imaging findings, pathological evidence, risk factors, and treatment considerations to provide balanced, clinically grounded assessments across the full breast care pathway. Now, apply your expertise to the following task: Generate a descriptive caption for the provided breast ultrasound image(s). You MUST explicitly ground each mentioned finding with a 2D bounding box [xmin, ymin, xmax, ymax].

\textbf{User.} \texttt{<image>}\\
Locate any abnormalities in this breast ultrasound and provide a detailed description of their morphological features.

\textbf{Reference output.} Standard Ultrasound (B-Mode) reveals \texttt{<ref-object><bbox>}. This finding demonstrates microcalcifications of suspicious morphology and is categorized as BI-RADS 4A, prompting a recall for biopsy or further evaluation. Despite benign morphological features, the presence of suspicious calcifications elevates concern, though the overall prognosis remains favorable with routine follow-up.

\textbf{Comment.} The expected output interleaves a 2D bounding box with a narrative description, exercising both spatial localisation (scored by IoU in \cref{tab:caption_report}) and BI-RADS-level reasoning (scored by the weighted text metrics).
\end{promptbox}
\caption{Case 3: lesion-grounded captioning on a BUS expert-cropped frame from BUS-CoT/BUS-Expert.}
\label{fig:case3}
\end{figure}

\begin{figure}[H]
\centering
\includegraphics[width=0.30\linewidth]{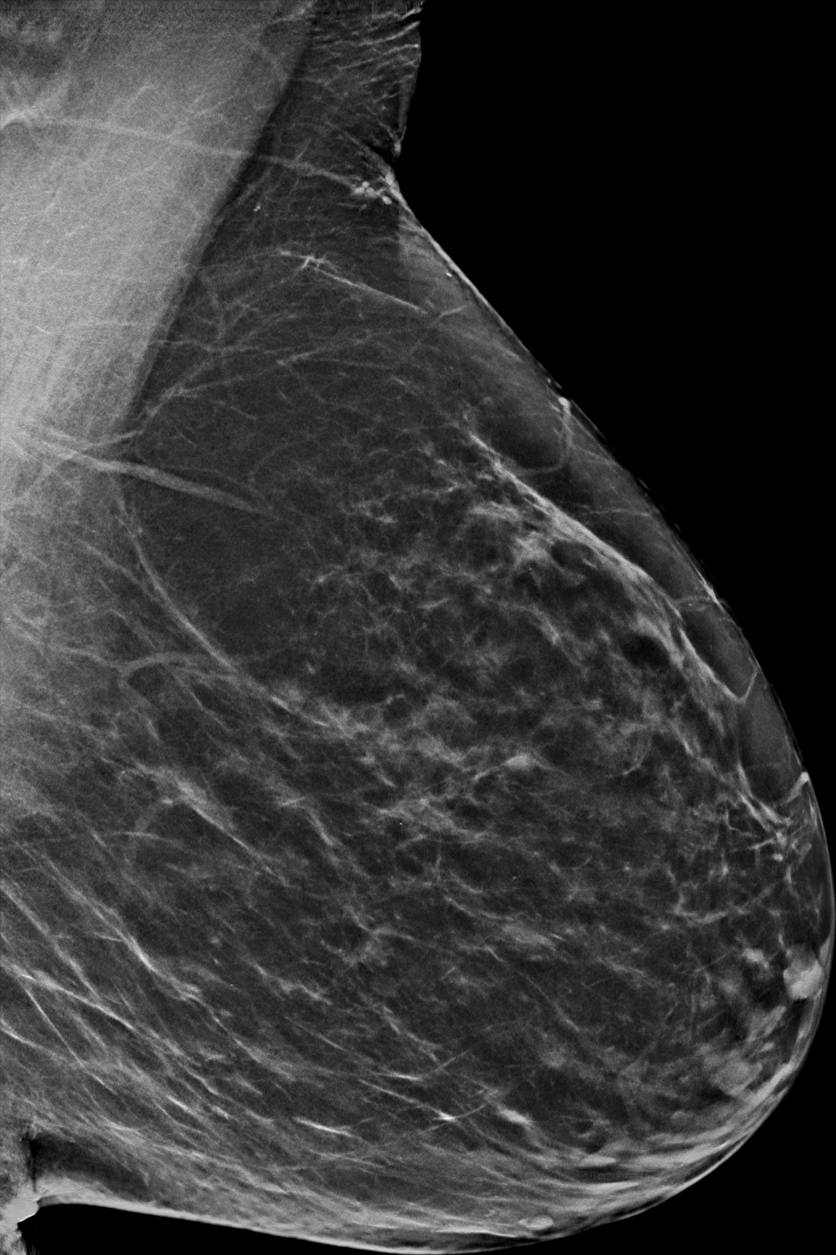}
\begin{promptbox}[{{Case 4 -- Mammography, Diagnosis-stage Closed VQA (BI-RADS)}}]{cyan}
\textbf{System.} You are an expert diagnostic breast radiologist. Your clinical focus is on definitive lesion characterization and differential diagnosis. You rigorously analyze imaging features to separate benign from malignant entities, providing conclusive assessments that directly guide patient management. Now, apply your expertise to the following task: Answer the multiple-choice question based on the provided mammogram image(s). Respond ONLY with the exact option chosen.

\textbf{User.} \texttt{<image>$\times$8}\\
Can you provide the Bi-Rads category for this mammography exam?\\
Options: (A) Bi-Rads 3 \quad (B) Bi-Rads 6 \quad (C) Bi-Rads 0 \quad (D) Bi-Rads 1 \quad (E) Bi-Rads 2 \quad (F) Bi-Rads 4 \quad (G) Bi-Rads 5\\
Answer:

\textbf{Reference output.} (D) Bi-Rads 1

\textbf{Comment.} The eight-view input forces multi-image fusion within a single forward pass; the output also illustrates BreastStage's commitment to the full BI-RADS\,0--6 enum, including the assessment-incomplete category (0) and the established-malignancy category (6) that purely diagnostic benchmarks frequently omit.
\end{promptbox}
\caption{Case 4: BI-RADS classification on an 8-view mammographic exam from EMBED (one view shown).}
\label{fig:case4}
\end{figure}

\begin{figure}[H]
\centering
\begin{minipage}{0.30\linewidth}\centering\includegraphics[width=\linewidth]{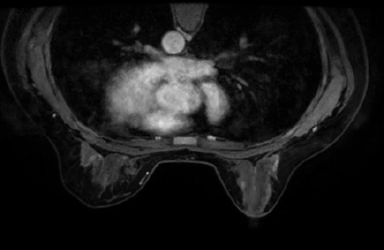}\\\footnotesize T1dyn\end{minipage}\hfill
\begin{minipage}{0.30\linewidth}\centering\includegraphics[width=\linewidth]{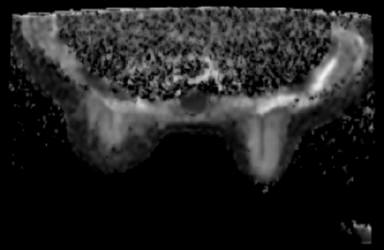}\\\footnotesize ADC\end{minipage}\hfill
\begin{minipage}{0.30\linewidth}\centering\includegraphics[width=\linewidth]{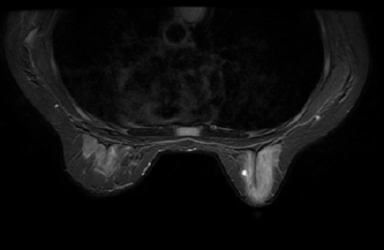}\\\footnotesize T2w\end{minipage}
\begin{promptbox}[{{Case 5 -- Multiparametric MRI, Screening-stage Report Generation}}]{purple}
\textbf{System.} You are an expert breast clinician. Your clinical focus is on comprehensive breast disease assessment and patient-centered clinical management. You integrate clinical presentation, imaging findings, pathological evidence, risk factors, and treatment considerations to provide balanced, clinically grounded assessments across the full breast care pathway. Now, apply your expertise to the following task: Generate a comprehensive clinical report based on the provided multi-sequence breast MRI.

\textbf{User.} Breast MRI sequences: ADC: \texttt{<video>}. DWI: \texttt{<video>}. T1: \texttt{<video>}. T1dyn: \texttt{<video>}. T2w: \texttt{<video>}.\\
Based on the provided breast MRI images, generate a comprehensive and professional MRI report detailing all relevant findings, diagnostic impression with BI-RADS classification, and appropriate clinical recommendations.

\textbf{Reference output.} \textbf{Breast MRI Report.}\\
\textbf{Findings:} The bilateral breasts are symmetric in size and contour, with heterogeneously dense (ACR Category C) fibroglandular tissue and mild background parenchymal enhancement. No architectural distortion or abnormal parenchymal distribution is identified. In the posteroinferior inner quadrant of the left breast, there is an irregular, spiculated mass measuring approximately 13\,mm $\times$ 11\,mm $\times$ 16\,mm. This lesion demonstrates marked early enhancement with washout kinetics and shows restricted diffusion on DWI/ADC sequences. No additional suspicious enhancing foci are seen in the left or right breast. Bilateral skin thickness is normal, nipples are symmetric without retraction, and there is no evidence of chest wall invasion. Axillary lymph nodes bilaterally appear morphologically benign without pathological enlargement. An incidental finding of a simple hepatic cyst is noted in the liver.\\
\textbf{Impression:} Irregular, markedly enhancing mass in the posteroinferior inner quadrant of the left breast with diffusion restriction, consistent with known malignancy. BI-RADS 6. Right breast shows no MRI evidence of malignancy. BI-RADS 1. Incidental simple hepatic cyst, likely benign.\\
\textbf{Recommendation:} Given the BI-RADS 6 classification, this finding reflects a known or biopsy-proven malignancy. Proceed with definitive surgical management or neoadjuvant therapy planning as per multidisciplinary consensus. No further diagnostic imaging of the breast is required for this lesion. The hepatic cyst requires no intervention or follow-up.

\textbf{Comment.} The report task simultaneously requires multi-sequence integration, structured section ordering (Findings $\to$ Impression $\to$ BI-RADS $\to$ Management), and quantitative descriptors that must be supported by the actual volumes; failures concentrate here in our \cref{tab:caption_report} numbers.
\end{promptbox}
\caption{Case 5: comprehensive multiparametric MRI report from FUDAN; mid-slice frames of three of the five paired sequences shown above.}
\label{fig:case5}
\end{figure}

\begin{figure}[H]
\centering
\begin{minipage}{0.22\linewidth}\centering\includegraphics[width=\linewidth]{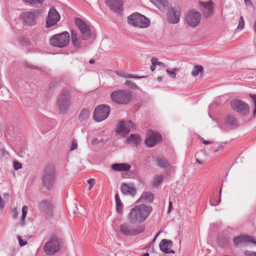}\\\footnotesize patch 1\end{minipage}\hfill
\begin{minipage}{0.22\linewidth}\centering\includegraphics[width=\linewidth]{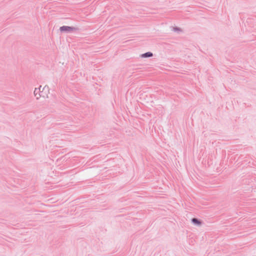}\\\footnotesize patch 9\end{minipage}\hfill
\begin{minipage}{0.22\linewidth}\centering\includegraphics[width=\linewidth]{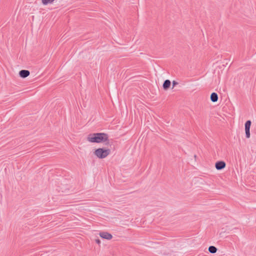}\\\footnotesize patch 17\end{minipage}
\begin{promptbox}[{{Case 6 -- Histopathology WSI, Pathology Report Generation}}]{red!60!black}
\textbf{System.} You are an expert breast clinician. Your clinical focus is on comprehensive breast disease assessment and patient-centered clinical management. You integrate clinical presentation, imaging findings, pathological evidence, risk factors, and treatment considerations to provide balanced, clinically grounded assessments across the full breast care pathway. Now, apply your expertise to the following task: Generate a concise descriptive caption for the provided breast histopathological image(s).

\textbf{User.} \texttt{<image>$\times$5024}\\
Please generate a comprehensive and professional pathology report based on the provided slide from Breast, including key elements such as: (1) the histopathological diagnosis, (2) tumor morphology and growth patterns, (3) extent of invasion and involved structures, (4) lymphovascular or perineural invasion, (5) status of surgical margins, (6) TNM staging if possible, and (7) any additional pathological findings including non-neoplastic tissue alterations.

\textbf{Reference output.} The pathology findings indicate a diagnosis of infiltrating ductal carcinoma located in the upper outer quadrant of the left breast. The tumor exhibits a moderately differentiated histologic grade, suggesting an intermediate level of aggression and atypia. There is grossly evident lesion upon inspection. Importantly, the surgical margins are uninvolved, indicating no residual tumor at the edges of the excised tissue. Lymph node analysis reveals that all ten axillary lymph nodes examined are negative for metastatic involvement, suggesting localized disease without regional spread. The Nottingham score, which typically helps in grading the tumor, could not be determined in this case. Overall, the pathological evaluation supports a moderately aggressive tumor confined to the primary site with no lymphatic dissemination.

\textbf{Comment.} On gigapixel inputs the GigaPixel branch must compress thousands of patch tokens down to $k{=}128$ before the LLM ever sees them; the section structure of the reference output (diagnosis $\to$ grade $\to$ margins $\to$ nodes $\to$ TNM) is exactly what the coverage selector is asked to preserve.
\end{promptbox}
\caption{Case 6: pathology report on a TCGA-BRCA whole-slide image at 20$\times$ magnification; 3 of the 5024 sampled patches shown above.}
\label{fig:case6}
\end{figure}

\paragraph{What the cases collectively show.}
The hard cases for BreastGPT cluster around three axes: (1)~fine-grained categorical boundaries when the schema admits adjacent values (BI-RADS\,3 vs.\ 4A, Nottingham grade II vs.\ III); (2)~quantitative predictions that should be supported by an explicit measurement but are inferred from the visual signal alone (ADC values, Ki-67 index); and (3)~molecular-biomarker inference (ER/PR/HER2 status) from H\&E morphology alone, without paired immunohistochemistry. These observations motivate future work on uncertainty-aware reporting, calibrated abstention on adjacent-category questions, and explicit integration of non-image clinical evidence at inference time.

\end{document}